\pdfoutput=1

\documentclass[11pt]{article}

\usepackage{ACL2023}

\usepackage{times}
\usepackage{latexsym}

\usepackage[T1]{fontenc}

\usepackage{graphicx}
\usepackage{float}
\usepackage{color, colortbl}
\usepackage{xcolor}
\definecolor{Gray}{gray}{0.9}
\usepackage{makecell}
\usepackage{lipsum}
\usepackage[most]{tcolorbox}
\usepackage{float}
\usepackage{verbatim} 
\usepackage{placeins}
\usepackage{epstopdf}
\usepackage{algorithm}
\usepackage{amsmath}
\usepackage{amssymb}
\usepackage{pifont}
\usepackage{multirow}
\usepackage{subcaption}
\usepackage[utf8]{inputenc}

\usepackage{microtype}

\usepackage{inconsolata}

\usepackage{pdfpages} 

\usepackage{hyperref}
\usepackage{xurl}

\title{\textit{SANSKRITI}: A Comprehensive Benchmark for Evaluating Language Models' Knowledge of Indian Culture}

\author{
    Arijit Maji$^1$, 
    Raghvendra Kumar$^1$, 
    Akash Ghosh$^1$, 
    Anushka$^2$,
    Sriparna Saha$^1$ \\
    $^1$Department of Computer Science and Engineering, Indian Institute of Technology Patna, India \\
    $^2$Department of Computer Science, Banasthali Vidyapeeth University, Rajasthan, India \\
    \small
    \begin{tabular}[t]{@{}c@{}}
    \texttt{\{arijit\_2311ai25,raghvendra\_2221cs27,akash\_2321cs19,sriparna\}@iitp.ac.in, guptaanushka024@gmail.com}
    \end{tabular}
}

\begin{document}

\maketitle
\begin{abstract}
Language models (LMs) are indispensable tools shaping modern workflows, but their global effectiveness depends on understanding local socio-cultural contexts. To address this, we introduce \textit{\textbf{SANSKRITI}}, a benchmark designed to evaluate language models' comprehension of India's rich cultural diversity. Comprising of 21,853 meticulously curated question-answer pairs spanning 28 states and 8 union territories, \textit{\textbf{SANSKRITI}} is the largest dataset for testing Indian cultural knowledge. It covers sixteen key attributes of Indian culture namely rituals and ceremonies, history, tourism, cuisine, dance and music, costume, language, art, festivals, religion, medicine, transport, sports, nightlife and personalities, providing a comprehensive representation of India’s cultural tapestry. We evaluate \textit{\textbf{SANSKRITI}} on leading  Large Language Models (LLMs), Indic Language Models (ILMs), and Small Language Models (SLMs), revealing significant disparities in their ability to handle culturally nuanced queries, with many models struggling in region-specific contexts. By offering an extensive, culturally rich, and diverse dataset, \textit{\textbf{SANSKRITI}} sets a new standard for assessing and improving the cultural understanding of LMs. 


\end{abstract}

\section{Introduction}
Language Models (LMs) have revolutionized the way we interact with technology, becoming indispensable tools in applications such as natural language understanding, content generation, knowledge discovery, and decision-making \citep{brown2020language,ghosh2024sights,devlin2019bert,ghosh2024healthalignsumm}. Models like Large Language Models (LLMs), Indic Language Models (ILMs), and Small Language Models (SLMs) are increasingly integrated into workflows across industries, enabling seamless communication and efficient problem-solving \citep{ouyang2022training} \footnote{In this study we define Large Language Model(LLM) as any language model of size more than or equal to 7 billion parameters, Small language models(SLM) are any that has less or equal to 5 billion parameters and Indic Language models (ILM) are any model that is developed inside India indigeneously}. These advancements, however, come with the critical challenge of ensuring that LMs cater to diverse populations by recognizing and reasoning about their linguistic and cultural contexts \citep{bender2021dangers,ghosh2025multilingual}.\par

While LMs excel in syntactic and semantic language understanding, their effectiveness depends significantly on their ability to comprehend cultural diversity \citep{lin2021understanding}. A model’s capacity to recognize cultural contexts is crucial for its application in education, governance, healthcare, and entertainment in culturally rich societies \citep{liang2022holistic,ghosh2024medsumm}. The inability to incorporate cultural nuances risks perpetuating biases, stereotypes, or inaccuracies, thereby alienating underrepresented communities. Models that can capture such diversity not only perform better but also foster inclusivity and equity in AI-driven solutions \citep{blodgett2020language}.\par

India’s unparalleled cultural diversity makes it an ideal testbed for evaluating the cultural competence of LMs. The country is home to 28 states and 8 union territories, each characterized by distinct traditions, festivals, cuisines, music, dance forms, architectural styles, and historical narratives \citep{sharma2019indian_culture}. For example, states like Rajasthan and Tamil Nadu exhibit unique artistic and culinary traditions, while festivals such as Diwali, Eid, and Onam highlight diverse religious and regional practices \citep{narayanan2020festivals,tripathi2019indian_culture}.\par

This cultural richness presents significant challenges for LMs. Existing benchmarks like TyDi QA \citep{joshi2020tydiqa} and XQUAD \citep{artetxe2020cross} primarily focus on multilingualism, overlooking cultural depth. Models trained on global datasets often fail to capture region-specific nuances, such as the significance of Kathakali performances in Kerala or Madhubani art in Bihar \citep{patel2023cultural_models}. This gap in understanding not only limits the utility of these models in culturally complex regions but also hinders their acceptance by local populations \citep{liang2022holistic}.

\textbf{Motivation for This Research :}  
The absence of culturally refined benchmarks, especially based on different nuanced attributes, inspired the creation of  \textit{\textbf{SANSKRITI}}, a dataset designed to evaluate the cultural understanding of LMs in the Indian context. Unlike existing datasets like \citet{seth2024dosa}, which are smaller and do not include all Indian states and union territories, \textit{\textbf{SANSKRITI}} focuses on cultural aspects across all 28 states and 8 union territories of India, providing a complete picture of the country’s rich cultural diversity. The motivation behind this research lies in bridging the gap between technological advancements in LMs and their real-world applicability in culturally diverse societies. 
We developed the \textit{\textbf{SANSKRITI}} dataset to benchmark cultural understanding in India, utilizing diverse data sources to cover key attributes like rituals and ceremonies, history, tourism, cuisine, dance and music, costume, language, art, festivals, religion, medicine, transport, sports, nightlife, and personalities. A team of 40 annotators curated 21,853 questions across four categories: association, country prediction, general awareness, and state prediction. 
\par

\textbf{ Research Questions:}  
This research aims to address the following questions:\par  
1. How do various language models (LLMs, SLMs, and ILMs) perform on the \textit{\textbf{SANSKRITI}} dataset, and what are the performance trends across these model categories?\par  
2. How do language models perform across different cultural attributes represented in the \textit{\textbf{SANSKRITI}} dataset? \par 
3. What are the performance trends of language models across different Indian states and union territories?  \par
4. How do language models perform across various question types in the \textit{\textbf{SANSKRITI}} dataset? \par  
\textbf{Contributions:}  
Our key contributions are summarized as follows:\par 

1. \textbf{Introduction of SANSKRITI}: We present \textit{\textbf{SANSKRITI}}, a dataset of 21853 manually curated and validated question-answer pairs, designed to evaluate the cultural competence of LMs in the Indian context.\par  
2. \textbf{Comprehensive Cultural Coverage}: \textit{\textbf{SANSKRITI}} encompasses 16 key cultural attributes spanning all Indian states and union territories.\par  
3. \textbf{Benchmarking Across Models}: We evaluate leading LLMs, ILMs, and SLMs on \textit{\textbf{SANSKRITI}}, identifying critical gaps in their ability to reason about culturally nuanced queries. \par
4. \textbf{Public Availability}: The \textit{\textbf{SANSKRITI}} dataset and benchmarking results are made publicly available to encourage research in culturally inclusive AI. The resources are available at:
\textbf{HuggingFace -} \url{https://huggingface.co/datasets/13ari/Sanskriti},  
\textbf{Google Drive - } \url{https://drive.google.com/drive/folders/1UEkhrcA3-aPQTjTwSIgBC8EbqSS8ntDn?usp=sharing}.\par

\section{Related Works}
\subsection{Culture in LLMs}
Recent studies have examined the sociocultural reasoning of large language models (LLMs), focusing on their adaptability to diverse cultural values and contexts. Researchers have used tools like the World Values Survey and Hofstede's dimensions to evaluate LLMs' understanding of social norms and human values \citep{johnson2022ghost,atari2023morality,masoud2023cultural}.  Other works have explored cultural artifacts, such as food and art, highlighting LLMs' knowledge but limited adaptability to user-specific scenarios \citep{seth-etal-2024-dosa,li2024culture}. Efforts to improve adaptability include enabling synthetic personas, but studies reveal gaps in understanding non-Western cultures and persistent biases \citep{alkhamissi2024investigating,durmus2023towards}. Fine-tuning approaches have been applied to instill social norms and enhance performance on culture-specific tasks like hate speech detection, though probing in regional languages often underperforms compared to monolingual English testing \citep{dwivedi2024exploring,shen2017style}.

\subsection{NLP for Indian Languages}
Recent advancements in large language models (LLMs) tailored for India focus on enhancing multilingual capabilities and addressing the country’s linguistic diversity. Notable contributions include IndicGenBench, a multilingual benchmark for 29 Indic languages to evaluate generation capabilities across scripts and language families, promoting research on underrepresented languages 
\citep{singh2024indicgenbench}. MuRIL (short for Multilingual Representations for Indian Languages), a language model specifically designed for Indian languages, effectively handles transliterated and code-mixed text commonly found in informal settings \citep{khanuja2021muril}. The LoFTI benchmark (acronym for Localization and Factuality Transfer to Indian Locales) evaluates LLMs’ ability to provide localized and factual responses in Indian contexts \citep{simon2024lofti}. Additionally, studies on translation capabilities of LLMs between English and 22 Indian languages highlight advancements in multilingual tasks through in-context learning and fine-tuning \citep{patel2023cultural_models}.

\subsection{Collaborative Dataset Creation}
Participatory research involves those impacted by technology in its design and evaluation. While widely used in HCI, such methods are underutilized in NLP \cite{diddee2022six}. Datasets like DOSA (short for ``Dataset of Social Artifacts") \citet{seth2024dosa} and CVQA (acronym for ``Culturally-diverse Multilingual Visual Question Answering") \citet{romero2024cvqa} highlight the importance of collaboration, using local input to capture cultural artifacts and diverse insights. Games with a purpose (GWAP) have proven effective for eliciting implicit knowledge \citep{balayn2022ready,von2006verbosity}. Building on this, we use surveys and a flexible adaptation of the Taboo game to collect cultural artifacts and their significance \citep{stephenson2023culture}, enabling richer and more context-specific data.

\section{\textit{Development of \textit{\textbf{SANSKRITI}}}}

\subsection{Data Collection}

\textbf{Data Sources:} The creation of the \textit{\textbf{SANSKRITI}} benchmark was a meticulous, multi-phase endeavour designed to ensure comprehensive coverage and high-quality content. The dataset was constructed by sourcing data from six diverse and reliable platforms: Wikipedia\footnote{\url{https://www.wikipedia.org}}, Ritiriwaz\footnote{\url{https://www.ritiriwaz.com}}, Holidify\footnote{\url{https://www.holidify.com}}, Artsandculture\footnote{\url{https://artsandculture.google.com}}, and Timesofindia\footnote{\url{https://timesofindia.indiatimes.com}}. These sources were carefully selected to capture the rich tapestry of India's cultural heritage with authenticity and depth.

\textbf{Wikipedia:} A well-documented and detailed resource, Wikipedia served as the cornerstone for verified knowledge across diverse domains such as history, arts, languages, and traditions (see appendix for the full domain list). 

\textbf{Ritiriwaz:} Ritiriwaz contributed region-specific insights into Indian customs, rituals, and traditions, providing a valuable perspective on unique practices.

\textbf{Holidify:} Focused on travel, Holidify provided information on festivals, cuisines, and landmarks, effectively linking cultural elements to their geographical origins.

\textbf{Arts and Culture:} Arts and culture enhanced the dataset with visually engaging explorations of art, music, and heritage, complementing textual data with multimedia depth.

\textbf{Times of India:} As a leading news outlet, Times of India provided a contemporary angle, covering cultural events, trends, and regional highlights.

\textbf{Data Organization:} These sources collectively provided a robust repository of historical, traditional, and modern cultural insights. The information was systematically organized into a structured format (\texttt{{``state name":{``attribute":
``scrapped data related to the attribute and state"}}}) to facilitate efficient processing and analysis, ensuring a balanced and authentic representation of India’s cultural diversity.

\subsection{Annotation Process} 

\textbf{Questions Formation:} After collecting the raw data, we curated a structured list of cultural categories—such as arts, festivals, cuisines, music and dance, languages, histories, and traditions (full list in the appendix \ref{complete-list})—inspired by \citet{seth-etal-2024-dosa,romero2024cvqa}. These categories were chosen to capture the multifaceted essence of India’s cultural heritage, ensuring comprehensive coverage and balanced representation. The questions in this benchmark are designed as \textbf{fact-based} multiple-choice queries.

\textbf{Cultural Dimensions:} By organizing the dataset around these key dimensions, we enable detailed analyses and facilitate applications that require a deep understanding of India's diverse cultural landscape, significantly enhancing the dataset's utility.

\textbf{Team Structure:} To generate a comprehensive and high-quality question dataset, we employed a team of forty members, divided into four specialized sub-teams of ten members each. Each sub-team focused on one of the following question types:  


\textbullet \textbf{Association Prediction:} Identify the culture referenced in a statement. 

\textbullet \textbf{Country Prediction:} Determine the country based on the statement.

\textbullet \textbf{General Awareness Prediction:} Formulate factual questions.

\textbullet \textbf{State Prediction:} Identify the specific Indian state referenced in a statement.

\textbf{Quality Assurance:} This focused division of labour ensured both efficiency and thoroughness. A total of 21,853 questions were created across these categories, with each sub-team dedicated to its assigned type to maintain quality and minimize redundancy.

\textbf{Guidelines:} To ensure consistency and quality, clear and detailed guidelines were established, outlining definitions for each question type, attributes, sample questions, and cultural considerations. Teams were instructed to prioritize accuracy, diversity, and relevance while avoiding ambiguity or excessive complexity. Outputs from each sub-team underwent cross-validation by another sub-team, fostering collaboration to resolve ambiguities or inaccuracies. A final review ensured adherence to all guidelines.

\textbf{Team Training and Distribution:} Team members, selected for their expertise in linguistics, Indian culture, or related fields, underwent a brief training session on the dataset’s objectives, question types, and cultural sensitivity to maintain a unified approach. Questions were evenly distributed across four types, with minor variations emerging naturally during creation. Redundancy was mitigated by cross-checking for similar content, ensuring diverse coverage of cultural elements.

\textbf{Cultural Sensitivity and Ethical Considerations:} To promote equitable representation, cultural references spanned all Indian states and union territories, with a particular focus on less-represented regions and unique artifacts. Ethical guidelines emphasized cultural sensitivity, avoided stereotypes, and encouraged respectful and inclusive question framing. The cross-validation process further identified and addressed potential biases or misrepresentations, ensuring a balanced and culturally respectful dataset. 

\textbf{Compensation Structure:} Lastly, each annotator was compensated for their contributions. The payment structure was designed to reflect the effort involved in each task: annotators were paid 1.20 USD for creating every 10 questions and 0.60 USD for verifying every 10 questions.

\begin{figure}[!htbp]
\centerline{\includegraphics[width=0.95\columnwidth]{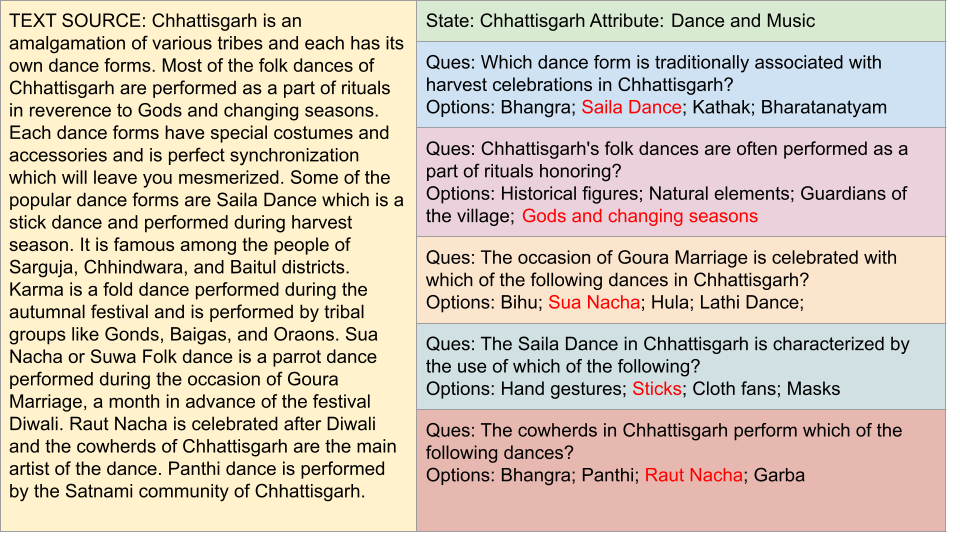}}
\caption{Illustration showcasing crafting questions on Chhattisgarh's dance and music Heritage.}
\label{sample}
\end{figure}

\textbf{Excerpt of Process:} Figure \ref{sample} illustrates an excerpt of the process used to create the questions shown on the right-hand side of the image, derived from the source text on the left. This process involved identifying key details, themes, and cultural elements described in the passage about Chhattisgarh's dance forms. Each question was carefully crafted to test specific factual knowledge from the text. For example, the association of the Saila Dance with the harvest season, the ritualistic purpose of folk dances, and the connection of Sua Nacha to Goura Marriage were directly drawn from explicit references in the source. Plausible distractors were included for each question, while the correct answer, highlighted in red, closely adhered to the source text. The question formulation prioritized clarity and relevance, aligning each item with distinct phrases or concepts (e.g., ``stick dance," ``rituals honouring gods and seasons," and ``cowherds performing Raut Nacha").

\begin{figure}[!htbp]
\centerline{\includegraphics[width=0.95\columnwidth]{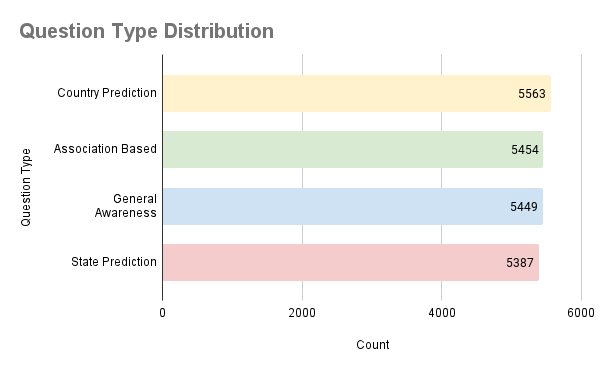}}
\caption{Distribution of Question Types Showing Balanced Representation Across Categories}
\label{ques-types}
\end{figure}

\subsection{Statistical details of \textit{\textbf{SANSKRITI}}}

\begin{figure}[!htbp]
\centerline{\includegraphics[width=0.95\columnwidth]{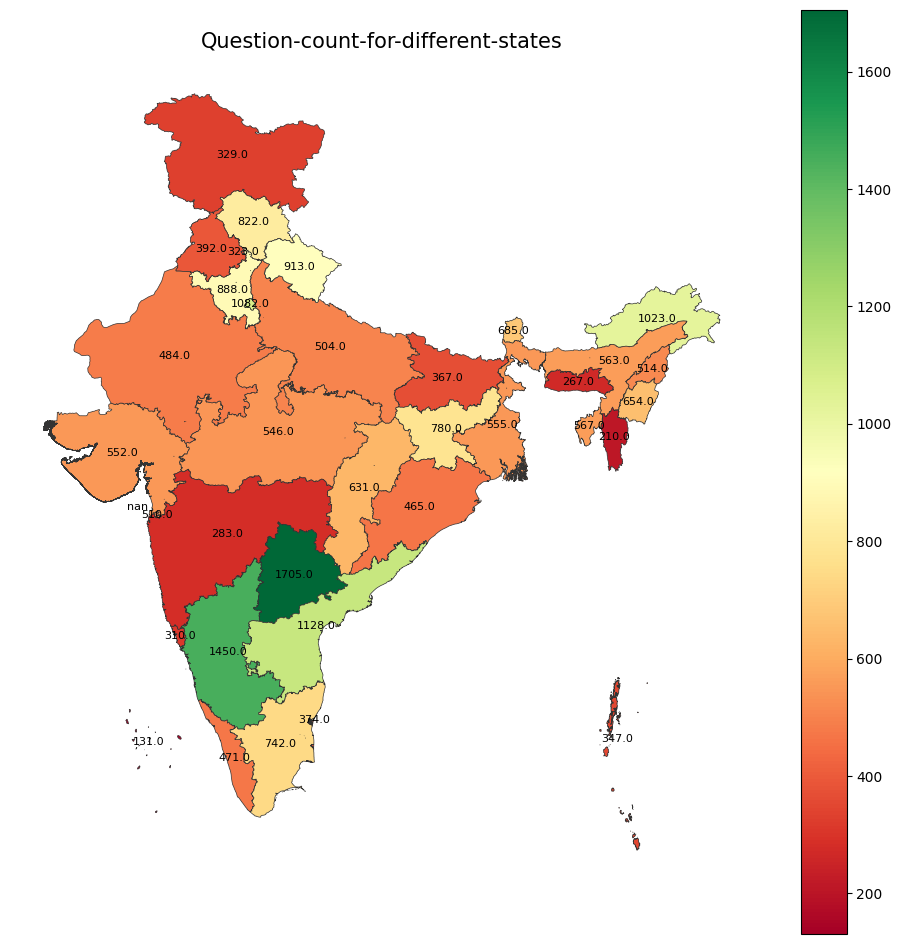}}
\caption{A map of India illustrating the distribution of questions across all states and union territories.}
\label{india-map}
\end{figure}

\begin{figure*}[!htbp]
\centerline{\includegraphics[width=0.95\textwidth]{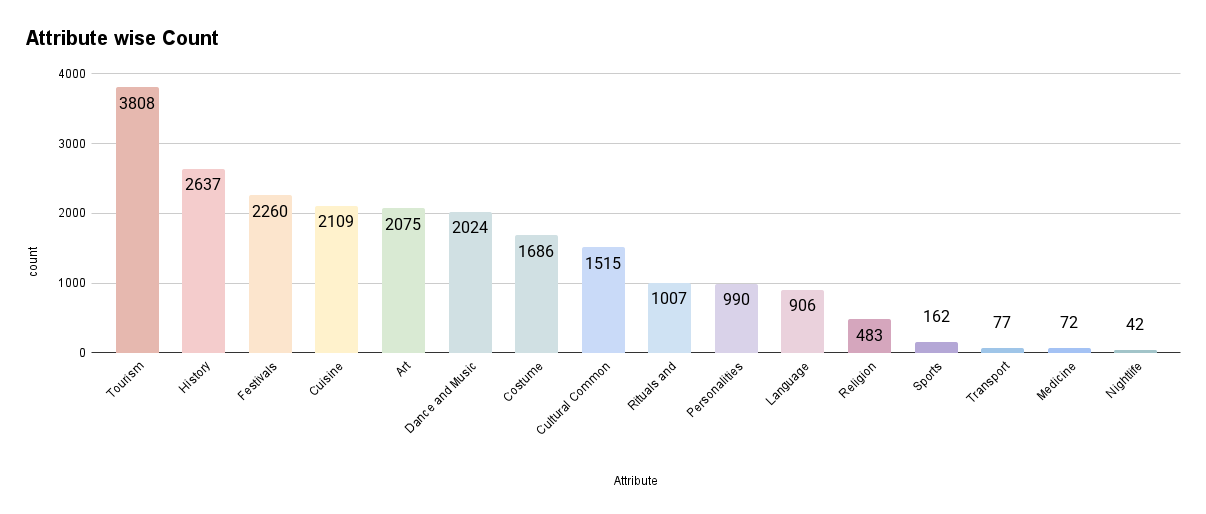}}
\caption{Distribution of Question Types across various Attributes.}
\label{attributes}
\end{figure*}

Our multidimensional dataset is analyzed through three key visualizations: a bar chart of question types (Figure~\ref{ques-types}), a state-wise India map (Figure~\ref{india-map}), and a bar chart of attribute-wise question counts (Figure~\ref{attributes}), offering a concise overview of its structure, diversity, and coverage.

\textbf{Distribution of Questions across States and Union Territories:} As shown in Figure \ref{india-map}, the map visualizes the distribution of question counts across different Indian states using a colour gradient where green represents higher counts and red represents lower counts. States like Telangana, Karnataka, and Andhra Pradesh have relatively higher question counts, signifying high engagement or activity. Conversely, Jammu and Kashmir, Maharashtra, and smaller northeastern states like Tripura show lower counts, indicated by red and orange shades. The range spans 300 to over 800, with central and southern states exhibiting stronger performance, while northern and northeastern regions reflect mixed trends. We must highlight that we have treated Dadra, Nagar Haveli, Daman, and Diu as a single entity in our work. This decision was made because, when considered separately, each region had a relatively small number of questions, whereas combining them resulted in a more significant count.

\textbf{Distribution of Questions across the Question Categories:} The bar chart shown in Figure \ref{ques-types} illustrates the distribution of various question types by their counts. Among the four categories, ``Country Prediction" has the highest representation, closely followed by ``Association Based" and ``General Awareness," which are almost equal in count. ``State Prediction" has slightly fewer instances but remains comparable to the other categories. Overall, the chart shows a balanced distribution of question types, with no significant disparity among them, indicating a well-rounded approach to creating diverse question types in the dataset.

\textbf{Distribution of Questions across the Attributes:} The bar chart presented in Figure \ref{attributes}, shows the distribution of attributes based on their count. It reveals that ``Tourism", ``Festivals," and ``History" are the most frequently represented attributes, indicating a strong focus on cultural and historical themes. Other significant attributes include ``Art," ``Cuisine," and ``Dance and Music," showcasing an emphasis on elements reflecting regional heritage and traditions. ``Costume" and ``Cultural Common Sense" also have notable representation, highlighting their importance in the context of cultural identity. Attributes like ``Languages" and ``Rituals" are moderately represented, while ``Religion" has relatively lower prominence. The least represented attributes include "Medicine," ``Transport," ``Sports," and ``Nightlife," suggesting these are less explored or emphasized within the dataset. Overall, the chart prioritizes cultural and historical aspects over contemporary or niche topics.

\begin{figure}[!htbp]
\centerline{\includegraphics[width=0.95\columnwidth]{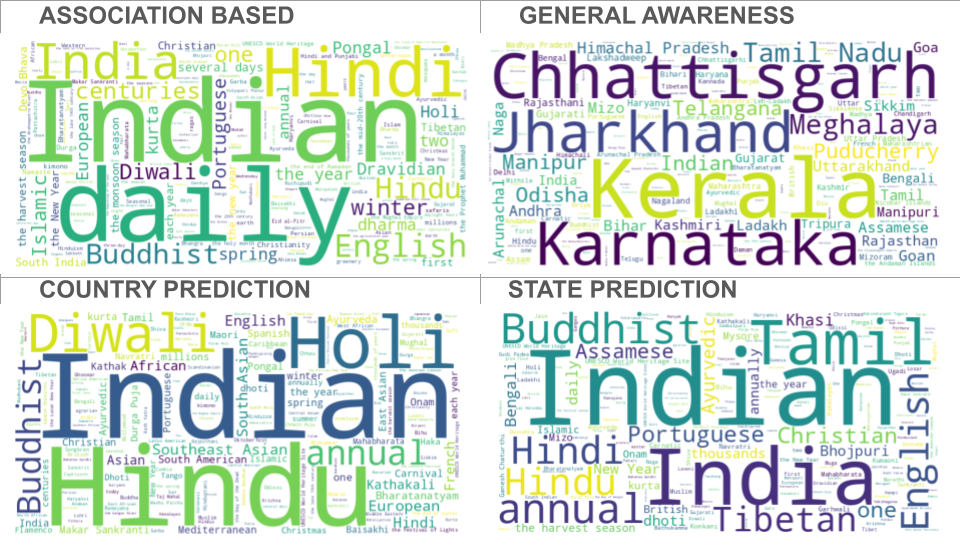}}
\caption{Word Cloud Representation of Question Categories}
\label{wordcloud}
\end{figure}

\textbf{Word Cloud Representation:} Last but not least, Figure \ref{wordcloud} presents word cloud visualizations for the four question types: Association Based, General Awareness, Country Prediction, and State Prediction. The Association Based category highlights key cultural and linguistic terms, while General Awareness emphasizes geographical knowledge through prominent state names. Country Prediction reflects national identity and traditions, whereas State Prediction focuses on regional and cultural themes. These word clouds collectively underscore the dataset's strong emphasis on cultural, religious, and geographical associations.

\section{Experimental  Section}

\subsection{Models}
To conduct a comprehensive evaluation of our benchmark, \textit{\textbf{SANSKRITI}}, we performed an extensive assessment across a diverse range of language models.  Our evaluation included prominent Large Language Models (LLMs), such as Mistral-7B-Instruct, LLAMA-3.1-70B-Instruct, Qwen-2.5-72B-Instruct, and Phi3-medium-4k-Instruct. Additionally, we tested several Small Language Models (SLMs), including Gemma-2B-Instruct, Qwen2-1.5B-Instruct, LLAMA-3.2-3B-Instruct, and SmolLM-1.7B-Instruct.  For Indic LLMs, we evaluated Navrasa-2.0 and  OpenHathi-Instruct. Furthermore, we incorporated the proprietary model GPT-4o into our evaluation for a holistic comparison.\footnote{In the work GPT and GPT-4o are used interchangeably. Both mean the same in this context.}

\begin{table*}[ht]
    \centering
    \scalebox{0.48}{ 
    \renewcommand{\arraystretch}{1.2}
    \begin{tabular}{|l|c|c|c|c|c|c|c|c|c|c|c|}
    \hline
    \textbf{QuestionType} & \textbf{Gemma-2-2b} & \textbf{Qwen2-1.5B} & \textbf{Llama-3.2-3B} & \textbf{SmolLM-1.7B} & \textbf{Navarasa-2.0} & \textbf{OpenHathi-7B} & \textbf{Phi-3-medium-4k} & \textbf{Mistral-7B} & \textbf{Llama-3.1-70B} & \textbf{Qwen2.5-72B} & \textbf{GPT-4o}  \\ \hline
    \textbf{Association Prediction} & 0.53 & 0.75 & 0.43 & 0.18 & 0.26 & 0.3 & 0.71 & 0.58 & 0.85 & 0.80 & 0.82 \\ \hline
    \textbf{State Prediction} & 0.40 & 0.51 & 0.50 & 0.18 & 0.28 & 0.2 & 0.7 & 0.61 & 0.78 & 0.72 & 0.80 \\ \hline
    \textbf{GK Prediction} & 0.20 & 0.82 & 0.79 & 0.16 & 0.56 & 0.35 & 0.85 & 0.81 & 0.93 & 0.94 &  0.96 \\ \hline
    \textbf{Country Prediction} & 0.81 & 0.9 & 0.36 & 0.13 & 0.5 & 0.43 & 0.84 & 0.8 & 0.88 & 0.92 & 0.93 \\ \hline
    \end{tabular}
    }
    \caption{Performance comparison of various LLMs and SLMs across tasks including Association Prediction, State Prediction, GK Prediction, and Country Prediction. The table presents the accuracy achieved by each model.}
    \label{tab:llm_task_comparison}
\end{table*}

\begin{figure*}[hbt!]
    \centering
    \begin{subfigure}[b]{0.48\textwidth}
        \centering
        \includegraphics[height=7cm, width=7cm]{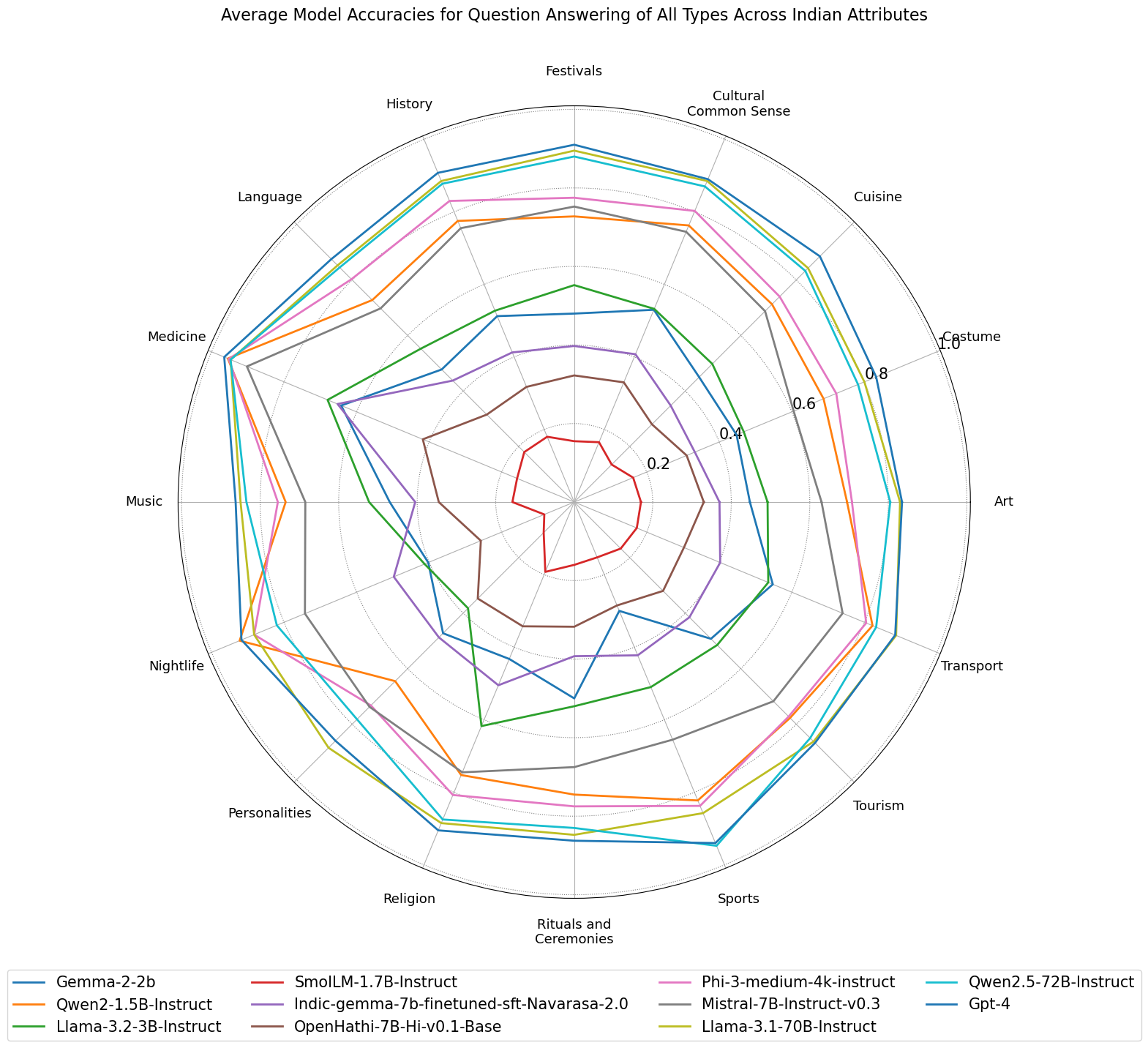}
        \caption{Classification based on attributes}
        \label{fig:metadata_pipeline}
    \end{subfigure}
    \hfill
    \begin{subfigure}[b]{0.48\textwidth}
        \centering
        \includegraphics[height=7cm, width=7cm]{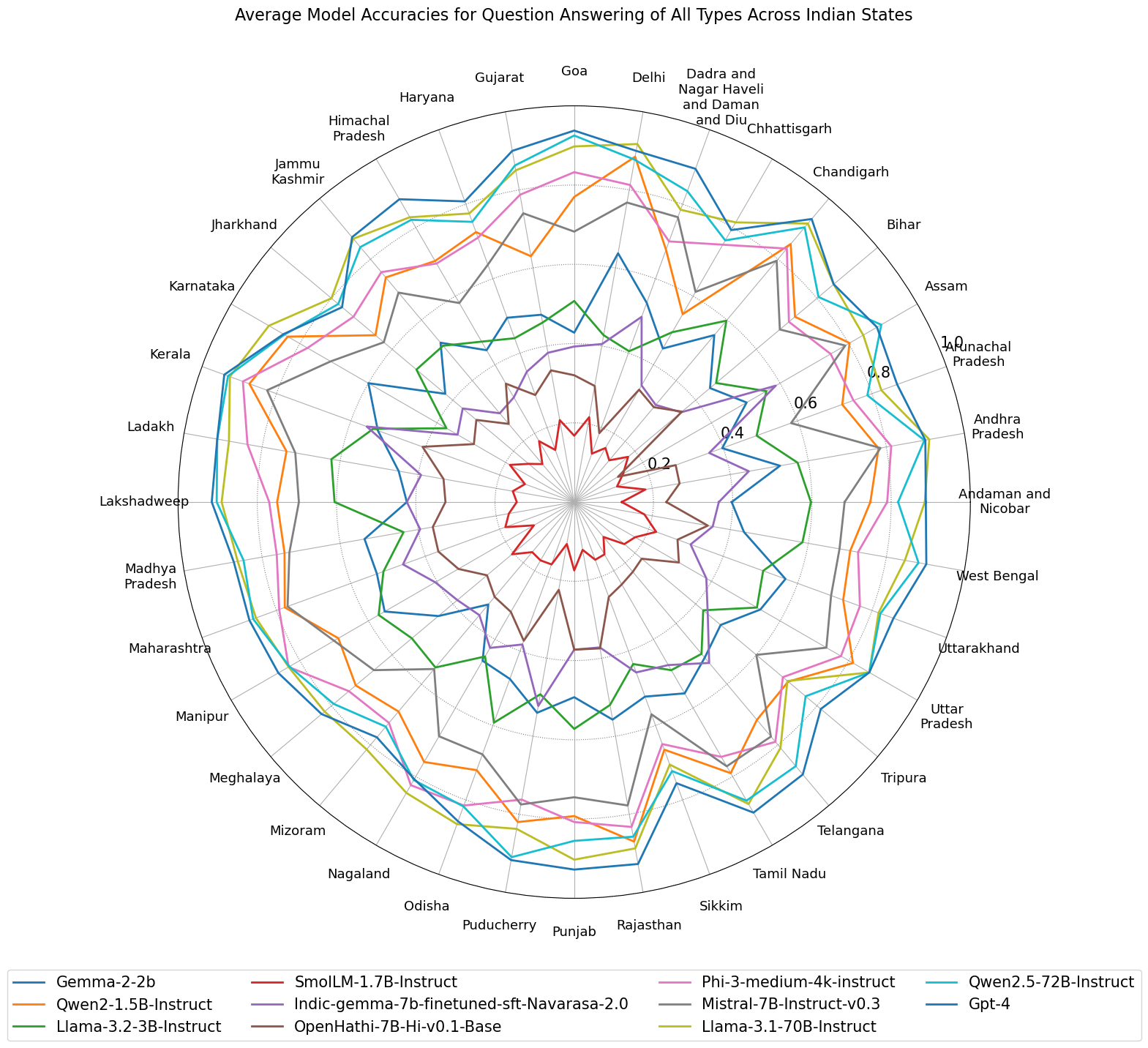}
        \caption{Classification based on states}
        \label{fig:infographic_pipeline}
    \end{subfigure}
    \caption{Average accuracy of language models on the \textit{\textbf{SANSKRITI}} dataset classified on the basis of attributes and states} 
    \label{fig:data_gen_pipeline}
\end{figure*}

\begin{table}[ht]
    \centering
    \scalebox{0.6}{ 
    \renewcommand{\arraystretch}{1.5}
    \begin{tabular}{|l|c|}
    \hline
    \textbf{Model} & \textbf{Attribute Score} \\ \hline
    Gemma-2-2b & 0.48 \\ \hline
    Qwen2-1.5B-Instruct & 0.74 \\ \hline
    Llama-3.2-3B-Instruct & 0.52 \\ \hline
    SmolLM-1.7B-Instruct & 0.16 \\ \hline
    Navarasa-2.0 & 0.40 \\ \hline
    OpenHathi-7B-Instruct & 0.32 \\ \hline
    Phi-3-medium-4k-Instruct & 0.77 \\ \hline
    Mistral-7B-Instruct-v0.3 & 0.70 \\ \hline
    Llama-3.1-70B-Instruct & 0.86 \\ \hline
    Qwen2.5-72B-Instruct & 0.84 \\ \hline
    GPT-4o & 0.87 \\ \hline
    \end{tabular}
    }
    \caption{Average accuracy of various language models on the entire \textit{\textbf{SANSKRITI}} dataset.}
    \label{tab:model_score}
\end{table}

\subsection{Evaluation Setup}

We conducted a zero-shot evaluation of cultural aspect-related questions, categorizing them into four main groups: \textbf{1. Country Prediction} : Prediction of the country based on cultural aspects, \textbf{2. State Prediction}: Prediction of the state in India is based on cultural aspects. 
\textbf{3. General Knowledge}: General awareness questions based on cultural attributes.  \textbf{4. Association-based Questions}  
All questions followed a \textbf{Multiple-Choice Question (MCQ)} format, with options (A, B, C, D).  The MCQs used to evaluate the cultural knowledge of the models are as follows: \textbf{1. Country Prediction:} (e.g., “Location: Unknown. Question: Which country is associated with {cultural\_aspect}? Options: {options} Short Answer:”). \textbf{2. State Prediction :} (e.g., “Location: India. Question: Which Indian state is known for {cultural\_aspect}? Options: {options} Short Answer:”). \textbf{3. General Knowledge :} (e.g., “Question: {general\_cultural\_question}? Options: {options} Short Answer:”). \textbf{4. Association-Based Questions:} (e.g., “Question: The {cultural\_entity} is most closely associated with which {cultural\_context}? Options: {options} Short Answer:”). For evaluation, we used \textbf{accuracy} as the sole metric. During inference, open-source models were run using 16-bit floating-point precision with \textbf{greedy decoding}, while proprietary models were accessed via their respective APIs. The models produced output probabilities for each option, and the option with the highest probability was selected as the prediction. This approach ensured consistency across all evaluations.

\section{Discussion on Results}

\subsection{Main Results}
 The overall performance on the \textit{\textbf{SANSKRITI}} dataset across various LLMs, SLMs, and ILMs is presented in Table \ref{tab:llm_task_comparison}. 

\textbf{Performance of LLMs :}  GPT-4o demonstrated the best performance on the \textit{\textbf{SANSKRITI}} dataset. Among the open-source LLMs, \textbf{LLAMA-3.1B-70B-Instruct} achieved the highest performance with a score of 0.86, followed by \textbf{Phi-3-medium-4k-Instruct} and \textbf{Mistral-7B-Instruct}, both showed comparable performance scores of 0.77 and 0.7, respectively.\par 

\textbf{Performance of SLMs:}
Among SLMs, \textbf{Qwen2-1.5B-Instruct} emerged as the best-performing SLM. Interestingly, it is also almost comparable to a few notable LLMs in our evaluation, highlighting that some SLMs possess better domain-specific knowledge than even larger LLMs. The performances of \textbf{Llama-3.2-3B-Instruct} and \textbf{SmolLM-1.7B-Instruct} were comparable, while \textbf{Gemmma-2-2B} with a score of 0.48  delivered the weakest performance among the SLMs.\par

\textbf{Performance of ILM:} Among the ILMs evaluated, \textbf{OpenHathi} emerged as the best performer with a score of 0.32. Surprisingly, \textbf{Navrasa-2.0}, despite being an ILM  delivered the weakest performance across all the language models evaluated on our benchmark \textit{\textbf{SANSKRITI}} dataset.

\subsection{Performance across Cultural Attributes} To assess the fine-grained capabilities of LLMs, SLMs, and ILMs in addressing questions related to various attributes of Indian culture, we present their accuracy comparison across different attributes in Figure-\ref{fig:infographic_pipeline} a. Our analysis reveals that for certain attributes, such as religion, medicine, and cultural common sense, the performance of all language models was notably high. However, for attributes like costumes, cuisines, and art, the language models generally struggled. Additionally, some trends were specific to certain LLMs—for instance, the performance of \textbf{Gemma-2-2b} dropped significantly for sports compared to other LLMs.

\subsection{Performance across Different States}
Similarly, we conducted a fine-grained analysis of language models across questions from various states and union territories of India. A notable trend was that language models struggled with questions pertaining to North-Eastern states such as Sikkim, Arunachal Pradesh, and Tripura. Poor performance was also observed for states like Bihar and Jharkhand. Conversely, language models performed relatively well for states like Delhi and Maharashtra. An interesting observation was that states with globally recognized cities tended to yield better results compared to others.

\subsection{Performance across Question Types}
We also evaluated the performance across different question types in \textit{\textbf{SANSKRITI}}. The evaluation revealed several noteworthy trends. Across all language models, the highest accuracy was achieved on \textbf{GK}-based questions, whereas \textbf{State Prediction} questions recorded the lowest accuracy. Overall, \textbf{GPT-4.0} emerged as the top-performing model in our evaluation. However, for \textbf{Association-Based} questions, \textbf{LLAMA-3.1B-70B} outperformed all other models, showcasing its strengths in this specific category. Remarkably, \textbf{Qwen2-1.5B-Instruct}, despite its smaller size, performed almost at par with the state-of-the-art \textbf{GPT-4.0} on the \textbf{Country Prediction} task, highlighting its efficiency and potential.

\subsection{Error Analysis}

To evaluate the strengths and limitations of the best-performing model, GPT-4o, we present an error analysis by examining questions grouped into correctly and incorrectly answered sets. This analysis underscores the model's proficiency in leveraging strong semantic associations while shedding light on challenges such as limited contextual understanding, ambiguous options, and gaps in the representation of subtle knowledge within the training data.

\begin{figure}[!htbp]
\centerline{\includegraphics[width=0.95\columnwidth]{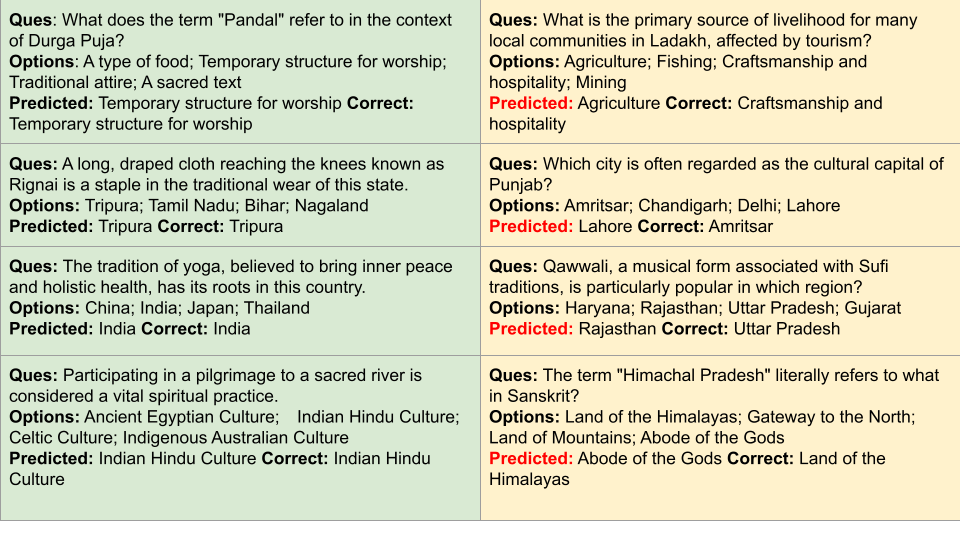}}
\caption{The LHS displays the correctly answered questions, while the RHS highlights the incorrectly answered ones.}
\label{error-analysis}
\end{figure}

As illustrated in Figure \ref{error-analysis}, the LHS showcases the model’s correct predictions, which likely arise from the strong semantic associations between keywords (e.g., ``Pandal", ``Yoga") and their contexts—areas well-represented in the training data. These questions are characterized by culturally specific, unambiguous topics with clear, non-overlapping options, enabling the model to effectively identify distinct patterns and contextual cues. Additionally, the clarity and relevance of the phrasing further aid in accurate classification, minimizing confusion.

Similarly, the RHS highlights the model’s incorrect predictions, which can be attributed to several factors. For instance, questions involving nuanced cultural or regional knowledge (e.g., primary livelihood in Ladakh, cultural capital of Punjab) require deeper contextual understanding, which the model may lack due to limited or biased training data. Ambiguity in certain options (e.g., ``Land of the Himalayas" vs. ``Abode of the Gods") further compounds the challenge, as similar terms can confuse the model. Additionally, less explicit semantic associations between keywords and correct answers, coupled with overlapping or less distinctive options, reduce the model's ability to identify the correct patterns effectively.

\section{Conclusion}
In this work, we introduced \textit{\textbf{SANSKRITI}}, a benchmark to evaluate language models' understanding of India's cultural diversity through over 21,000 curated question-answer pairs spanning 28 states and 8 union territories.  \textit{\textbf{SANSKRITI}} meets the need for culturally nuanced datasets by covering attributes like arts, festivals, and cuisines. Model evaluations revealed significant gaps, especially for certain states, highlighting biases likely rooted in training data limitations. \textit{\textbf{SANSKRITI}} dataset ensured quality, precision and cultural sensitivity, setting a new standard for inclusive AI research. By releasing  \textbf{SANSKRITI} publicly, we aim to advance the development of AI systems that are culturally aware and capable of serving diverse populations effectively. Looking forward, we plan to expand  \textbf{SANSKRITI} by adding more attributes to capture nuanced aspects of Indian culture, such as regional folklore and ecological heritage. Additionally, we aim to build a visual question-answering dataset and a multilingual version of \textbf{SANSKRITI}, enabling broader accessibility and applications across diverse linguistic and cultural contexts.

\section*{Limitations}
Although our study offers the most comprehensive evaluation of language models for Indian cultural knowledge, it has several notable limitations: \par
\textbf{1) Limited Scope of Cultural Attributes:}
This study covers only sixteen cultural attributes, which may not fully represent Indian culture. The question types also may not capture its full complexity. Future work will expand attributes and incorporate diverse QA formats, including True/False and adversarial questions.
    
\textbf{2) Lack of State-Specific Multilingual Queries:}  
    While our queries are multicultural, they do not currently include state-specific multilingual questions. Incorporating such queries is a key priority for the immediate extension of this work, as it will provide a more robust evaluation of language models across diverse linguistic contexts.
    
\textbf{3) Absence of Visual Question-Answering (VQA) Tasks:}  
    The dataset does not currently address visual question-answering tasks, which have become increasingly important with recent advancements in visual language models (VLMs). Future work will seek to incorporate VQA to further enhance the dataset's applicability and relevance.

\textbf{4) Limited Contextual Clarity:} Although we aimed to include attributes that are uniquely representative of Indian culture, certain examples in the benchmark may involve cultural elements that are somewhat ambiguous, as similar terms or practices can also be associated with other countries. In such cases, where the context is not detailed enough to clearly isolate India, we carefully design the multiple-choice questions so that the distractor options do not reference the attribute mentioned in the prompt. We then replace the original correct answer with the option that most closely aligns with it among the four choices.

\textbf{5) Lack of Diversity in Questions:} The questions in this benchmark are multiple-choice, fact-based questions that do not explicitly require reasoning or causal understanding to answer them.

\section*{Ethics Statement}

\textbf{Data Collection and Bias Mitigation:}  
The data used in the development of \textit{\textbf{SANSKRITI}} was sourced from publicly accessible platforms, as detailed in Section 3.1. While these platforms were carefully chosen to ensure authenticity and diversity, there remains a possibility of underrepresentation of certain cultural domains or regions. Despite its limitations, the \textit{\textbf{SANSKRITI}} dataset marks a significant milestone in establishing a standardized and inclusive benchmark for evaluating India's cultural knowledge. By encompassing all 28 states and 8 union territories, and featuring an extensive collection of 21,853 questions, it stands as the largest dataset dedicated to testing fine-grained cultural understanding of India. 

\textbf{Human Annotation and Cultural Sensitivity:}  
Human annotators played a key role in crafting and validating the dataset to ensure high quality and cultural accuracy. A diverse group of 40 annotators—selected for their expertise in Indian culture, linguistics, or related fields—included native and bilingual speakers from various Indian states. About 75\% were native speakers of at least one Indian language, and 80\% had lived over 15 years in regions where their primary language is spoken, ensuring strong cultural grounding. Annotators, aged 20 to 45, underwent training on dataset objectives, question categories, and cultural sensitivity. The annotation process was collaborative, with outputs cross-validated by a separate team to ensure consistency and mitigate bias. Ethical guidelines emphasized inclusivity and the avoidance of stereotypes, with any misrepresentations addressed during cross-validation to ensure respectful representation of India's diverse culture.
\section*{Acknowledgments}
Raghvendra Kumar gratefully acknowledges support from the Prime Minister's Research Fellowship (PMRF). Akash Ghosh expresses gratitude to the SERB POWER scheme (SPG/2021/003801), Department of Science and Technology, Government of India. Sriparna Saha acknowledges funding from the Technology Innovation Hub, Vishlesan I-Hub Foundation, IIT Patna (Project No. TIH/CSE/ASMO/05).

\bibliography{custom}

\begin{thebibliography}{35}
\expandafter\ifx\csname natexlab\endcsname\relax\def\natexlab#1{#1}\fi

\bibitem[{AlKhamissi et~al.(2024)AlKhamissi, ElNokrashy, AlKhamissi, and Diab}]{alkhamissi2024investigating}
Badr AlKhamissi, Muhammad ElNokrashy, Mai AlKhamissi, and Mona Diab. 2024.
\newblock Investigating cultural alignment of large language models.
\newblock \emph{arXiv preprint arXiv:2402.13231}.

\bibitem[{Artetxe et~al.(2020)}]{artetxe2020cross}
Mikel Artetxe et~al. 2020.
\newblock Xquad: A cross-lingual question answering dataset.
\newblock In \emph{EMNLP 2020}.

\bibitem[{Atari et~al.(2023)Atari, Haidt, Graham, Koleva, Stevens, and Dehghani}]{atari2023morality}
Mohammad Atari, Jonathan Haidt, Jesse Graham, Sena Koleva, Sean~T Stevens, and Morteza Dehghani. 2023.
\newblock Morality beyond the weird: How the nomological network of morality varies across cultures.
\newblock \emph{Journal of Personality and Social Psychology}.

\bibitem[{Balayn et~al.(2022)Balayn, He, Hu, Yang, and Gadiraju}]{balayn2022ready}
Agathe Balayn, Gaole He, Andrea Hu, Jie Yang, and Ujwal Gadiraju. 2022.
\newblock Ready player one! eliciting diverse knowledge using a configurable game.
\newblock In \emph{Proceedings of the ACM Web Conference 2022}, pages 1709--1719.

\bibitem[{Bender et~al.(2021)Bender, Gebru et~al.}]{bender2021dangers}
Emily~M. Bender, Timnit Gebru, et~al. 2021.
\newblock On the dangers of stochastic parrots: Can language models be too big?
\newblock \emph{FAccT 2021}, pages 610--623.

\bibitem[{Blodgett et~al.(2020)Blodgett, Barocas et~al.}]{blodgett2020language}
Su~Lin Blodgett, Solon Barocas, et~al. 2020.
\newblock Language (technology) is power: A critical survey of “bias” in nlp.
\newblock In \emph{ACL 2020}.

\bibitem[{Brown et~al.(2020)Brown, Mann, Ryder et~al.}]{brown2020language}
Tom Brown, Benjamin Mann, Nick Ryder, et~al. 2020.
\newblock Language models are few-shot learners.
\newblock \emph{Advances in neural information processing systems}, 33:1877--1901.

\bibitem[{Devlin et~al.(2019)Devlin, Chang, Lee, and Toutanova}]{devlin2019bert}
Jacob Devlin, Ming-Wei Chang, Kenton Lee, and Kristina Toutanova. 2019.
\newblock Bert: Pre-training of deep bidirectional transformers for language understanding.
\newblock In \emph{NAACL 2019}, pages 4171--4186.

\bibitem[{Diddee et~al.(2022)Diddee, Bali, Choudhury, and Mukhija}]{diddee2022six}
Harshita Diddee, Kalika Bali, Monojit Choudhury, and Namrata Mukhija. 2022.
\newblock The six conundrums of building and deploying language technologies for social good.
\newblock In \emph{Proceedings of the 5th ACM SIGCAS/SIGCHI Conference on Computing and Sustainable Societies}, pages 12--19.

\bibitem[{Durmus et~al.(2023)Durmus, Nyugen, Liao, Schiefer, Askell, Bakhtin, Chen, Hatfield-Dodds, Hernandez, Joseph et~al.}]{durmus2023towards}
Esin Durmus, Karina Nyugen, Thomas~I Liao, Nicholas Schiefer, Amanda Askell, Anton Bakhtin, Carol Chen, Zac Hatfield-Dodds, Danny Hernandez, Nicholas Joseph, et~al. 2023.
\newblock Towards measuring the representation of subjective global opinions in language models.
\newblock \emph{arXiv preprint arXiv:2306.16388}.

\bibitem[{Dwivedi and Patel(2024)}]{dwivedi2024exploring}
Sanjay~Kumar Dwivedi and Rahul Patel. 2024.
\newblock Exploring the intersections: Anthropological insights into studying language and culture.
\newblock \emph{State Institute of Education, Allahabad}, 30:171--182.

\bibitem[{Ghosh et~al.(2024{\natexlab{a}})Ghosh, Acharya, Jha, Saha, Gaudgaul, Majumdar, Chadha, Jain, Sinha, and Agarwal}]{ghosh2024medsumm}
Akash Ghosh, Arkadeep Acharya, Prince Jha, Sriparna Saha, Aniket Gaudgaul, Rajdeep Majumdar, Aman Chadha, Raghav Jain, Setu Sinha, and Shivani Agarwal. 2024{\natexlab{a}}.
\newblock Medsumm: A multimodal approach to summarizing code-mixed hindi-english clinical queries.
\newblock In \emph{European Conference on Information Retrieval}, pages 106--120. Springer.

\bibitem[{Ghosh et~al.(2024{\natexlab{b}})Ghosh, Acharya, Saha, Pandey, Raghu, and Sinha}]{ghosh2024healthalignsumm}
Akash Ghosh, Arkadeep Acharya, Sriparna Saha, Gaurav Pandey, Dinesh Raghu, and Setu Sinha. 2024{\natexlab{b}}.
\newblock Healthalignsumm: Utilizing alignment for multimodal summarization of code-mixed healthcare dialogues.
\newblock In \emph{Findings of the Association for Computational Linguistics: EMNLP 2024}, pages 11546--11560.

\bibitem[{Ghosh et~al.(2025)Ghosh, Datta, Saha, and Agarwal}]{ghosh2025multilingual}
Akash Ghosh, Debayan Datta, Sriparna Saha, and Chirag Agarwal. 2025.
\newblock The multilingual mind: A survey of multilingual reasoning in language models.
\newblock \emph{arXiv preprint arXiv:2502.09457}.

\bibitem[{Ghosh et~al.(2024{\natexlab{c}})Ghosh, Tomar, Tiwari, Saha, Salve, and Sinha}]{ghosh2024sights}
Akash Ghosh, Mohit Tomar, Abhisek Tiwari, Sriparna Saha, Jatin Salve, and Setu Sinha. 2024{\natexlab{c}}.
\newblock From sights to insights: Towards summarization of multimodal clinical documents.
\newblock In \emph{Proceedings of the 62nd Annual Meeting of the Association for Computational Linguistics (Volume 1: Long Papers)}, pages 13117--13129.

\bibitem[{Johnson et~al.(2022)Johnson, Pistilli, Men{\'e}dez-Gonz{\'a}lez, Duran, Panai, Kalpokiene, and Bertulfo}]{johnson2022ghost}
Rebecca~L Johnson, Giada Pistilli, Natalia Men{\'e}dez-Gonz{\'a}lez, Leslye Denisse~Dias Duran, Enrico Panai, Julija Kalpokiene, and Donald~Jay Bertulfo. 2022.
\newblock The ghost in the machine has an american accent: value conflict in gpt-3.
\newblock \emph{arXiv preprint arXiv:2203.07785}.

\bibitem[{Joshi et~al.(2020)}]{joshi2020tydiqa}
Pranav Joshi et~al. 2020.
\newblock Tydi qa: A benchmark for information-seeking question answering in typologically diverse languages.
\newblock In \emph{ACL 2020}.

\bibitem[{Khanuja et~al.(2021)Khanuja, Bansal, Mehtani, Khosla, Dey, Gopalan, Margam, Aggarwal, Nagipogu, Dave, Gupta, Gali, Subramanian, and Talukdar}]{khanuja2021muril}
Simran Khanuja, Diksha Bansal, Sarvesh Mehtani, Savya Khosla, Atreyee Dey, Balaji Gopalan, Dilip~Kumar Margam, Pooja Aggarwal, Rajiv~Teja Nagipogu, Shachi Dave, Shruti Gupta, Subhash Chandra~Bose Gali, Vish Subramanian, and Partha Talukdar. 2021.
\newblock \href {http://arxiv.org/abs/2103.10730} {Muril: Multilingual representations for indian languages}.

\bibitem[{Li et~al.(2024)Li, Jiang, Hwang, Kim, Santy, Sorensen, Lin, Dziri, Ren, and Choi}]{li2024culture}
Huihan Li, Liwei Jiang, Jena~D Hwang, Hyunwoo Kim, Sebastin Santy, Taylor Sorensen, Bill~Yuchen Lin, Nouha Dziri, Xiang Ren, and Yejin Choi. 2024.
\newblock Culture-gen: Revealing global cultural perception in language models through natural language prompting.
\newblock \emph{arXiv preprint arXiv:2404.10199}.

\bibitem[{Liang et~al.(2022)}]{liang2022holistic}
Percy Liang et~al. 2022.
\newblock Towards holistic evaluation of language models.
\newblock \emph{NeurIPS 2022}.

\bibitem[{Lin et~al.(2021)}]{lin2021understanding}
Bill~Yuchen Lin et~al. 2021.
\newblock Understanding cultural and gender biases in multilingual language models.
\newblock In \emph{NAACL 2021}.

\bibitem[{Masoud et~al.(2023)Masoud, Liu, Ferianc, Treleaven, and Rodrigues}]{masoud2023cultural}
Reem~I Masoud, Ziquan Liu, Martin Ferianc, Philip Treleaven, and Miguel Rodrigues. 2023.
\newblock Cultural alignment in large language models: An explanatory analysis based on hofstede's cultural dimensions.
\newblock \emph{arXiv preprint arXiv:2309.12342}.

\bibitem[{Narayanan(2020)}]{narayanan2020festivals}
Vishnu Narayanan. 2020.
\newblock Festivals of india: Cultural and religious significance.
\newblock \emph{Journal of Indian Traditions}.

\bibitem[{Ouyang et~al.(2022)Ouyang, Wu, Jiang et~al.}]{ouyang2022training}
Long Ouyang, Jeff Wu, Xu~Jiang, et~al. 2022.
\newblock Training language models to follow instructions with human feedback.
\newblock \emph{arXiv preprint arXiv:2203.02155}.

\bibitem[{Patel et~al.(2023)}]{patel2023cultural_models}
Ramesh Patel et~al. 2023.
\newblock Evaluating cultural understanding in multilingual models.
\newblock \emph{arXiv preprint arXiv:2302.12345}.

\bibitem[{Romero et~al.(2024)Romero, Lyu, Wibowo, Lynn, Hamed, Kishore, Mandal, Dragonetti, Abzaliev, Tonja et~al.}]{romero2024cvqa}
David Romero, Chenyang Lyu, Haryo~Akbarianto Wibowo, Teresa Lynn, Injy Hamed, Aditya~Nanda Kishore, Aishik Mandal, Alina Dragonetti, Artem Abzaliev, Atnafu~Lambebo Tonja, et~al. 2024.
\newblock Cvqa: Culturally-diverse multilingual visual question answering benchmark.
\newblock \emph{arXiv preprint arXiv:2406.05967}.

\bibitem[{Seth et~al.(2024{\natexlab{a}})Seth, Ahuja, Bali, and Sitaram}]{seth2024dosa}
Agrima Seth, Sanchit Ahuja, Kalika Bali, and Sunayana Sitaram. 2024{\natexlab{a}}.
\newblock Dosa: A dataset of social artifacts from different indian geographical subcultures.
\newblock \emph{arXiv preprint arXiv:2403.14651}.

\bibitem[{Seth et~al.(2024{\natexlab{b}})Seth, Ahuja, Bali, and Sitaram}]{seth-etal-2024-dosa}
Agrima Seth, Sanchit Ahuja, Kalika Bali, and Sunayana Sitaram. 2024{\natexlab{b}}.
\newblock \href {https://aclanthology.org/2024.lrec-main.474} {{DOSA}: A dataset of social artifacts from different {I}ndian geographical subcultures}.
\newblock In \emph{Proceedings of the 2024 Joint International Conference on Computational Linguistics, Language Resources and Evaluation (LREC-COLING 2024)}, pages 5323--5337, Torino, Italia. ELRA and ICCL.

\bibitem[{Sharma(2019)}]{sharma2019indian_culture}
Meera Sharma. 2019.
\newblock \emph{Indian Culture and Heritage: A Comprehensive Guide}.
\newblock Publisher XYZ.

\bibitem[{Shen et~al.(2017)Shen, Lei, Barzilay, and Jaakkola}]{shen2017style}
Tianxiao Shen, Tao Lei, Regina Barzilay, and Tommi Jaakkola. 2017.
\newblock Style transfer from non-parallel text by cross-alignment.
\newblock \emph{Advances in neural information processing systems}, 30.

\bibitem[{Simon et~al.(2024)Simon, Mondal, Singhania, Sen, and Jyothi}]{simon2024lofti}
Sona~Elza Simon, Soumen~Kumar Mondal, Abhishek Singhania, Sayambhu Sen, and Preethi Jyothi. 2024.
\newblock Lofti: Localization and factuality transfer to indian locales.
\newblock \emph{arXiv preprint arXiv:2407.11833}.

\bibitem[{Singh et~al.(2024)Singh, Gupta, Bharadwaj, Tewari, and Talukdar}]{singh2024indicgenbench}
Harman Singh, Nitish Gupta, Shikhar Bharadwaj, Dinesh Tewari, and Partha Talukdar. 2024.
\newblock Indicgenbench: A multilingual benchmark to evaluate generation capabilities of llms on indic languages.
\newblock \emph{arXiv preprint arXiv:2404.16816}.

\bibitem[{Stephenson(2023)}]{stephenson2023culture}
Janet Stephenson. 2023.
\newblock \emph{Culture and Sustainability: exploring stability and transformation with the cultures framework}.
\newblock Springer Nature.

\bibitem[{Tripathi(2019)}]{tripathi2019indian_culture}
Ramesh Tripathi. 2019.
\newblock Cultural diversity of india.
\newblock \emph{Indian Historical Review}.

\bibitem[{Von~Ahn et~al.(2006)Von~Ahn, Kedia, and Blum}]{von2006verbosity}
Luis Von~Ahn, Mihir Kedia, and Manuel Blum. 2006.
\newblock Verbosity: a game for collecting common-sense facts.
\newblock In \emph{Proceedings of the SIGCHI conference on Human Factors in computing systems}, pages 75--78.

\end{thebibliography}
\bibliographystyle{acl_natbib}

\section{Appendix}

\subsection{Complete List of Attributes and their definitions}\label{complete-list}
\begin{itemize}
    \item \textbf{Rituals and Ceremonies:} Traditions and practices performed during cultural or religious events.
    \item \textbf{Cultural Common Sense:} Shared knowledge and social norms unique to a culture.
    \item \textbf{History:} Chronological account of significant events and developments from the past.
    \item \textbf{Tourism:} Exploration of culturally or historically significant destinations.
    \item \textbf{Cuisine:} Traditional foods, cooking styles, and regional delicacies.
    \item \textbf{Dance and Music:} Artistic expressions through movement and sound representing cultural identity.
    \item \textbf{Costume:} Traditional attire and accessories specific to a region or culture.
    \item \textbf{Language:} Spoken and written forms of communication unique to a culture or region.
    \item \textbf{Art:} Visual and creative works, including paintings, sculptures, and crafts.
    \item \textbf{Festivals:} Celebrations marked by cultural or religious significance.
    \item \textbf{Religion:} Belief systems and practices centred around faith and spirituality.
    \item \textbf{Medicine:} Traditional healing practices and medical knowledge of a culture.
    \item \textbf{Transport:} Modes of travel and transportation characteristic of a region.
    \item \textbf{Sports:} Physical activities and games enjoyed for competition or recreation.
    \item \textbf{Nightlife:} Cultural activities and entertainment available during the evening.
    \item \textbf{Personalities:} Notable individuals who have significantly influenced cultural or historical narratives.
\end{itemize}

\subsection{Word Clouds based on questions classified  on  different  cultural attributes and Results on different kind of questions}



\begin{figure*}
    \centering
    \includegraphics[width=0.95\linewidth]{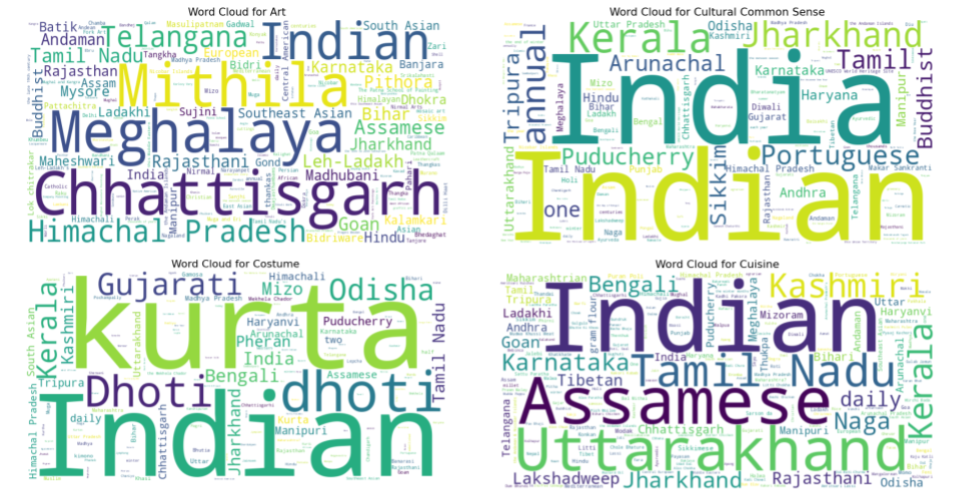}
    \caption{Word Cloud Representation based on attributes like arts, common sense, costumes, and cuisines.}
    \label{Word_Cloud_Representation_1}
\end{figure*}

\begin{figure*}
    \centering
    \includegraphics[width=0.95\linewidth]{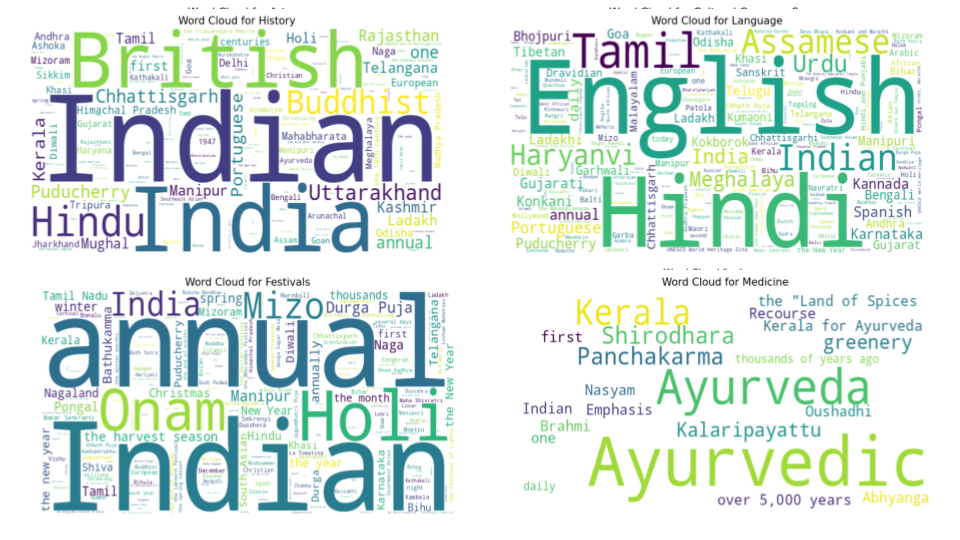}
    \caption{Word Cloud Representation based on attributes like music, nightlife, personalities and religion.}
    \label{Word_Cloud_Representation_2}
\end{figure*}

\begin{figure*}
    \centering
    \includegraphics[width=0.95\linewidth]{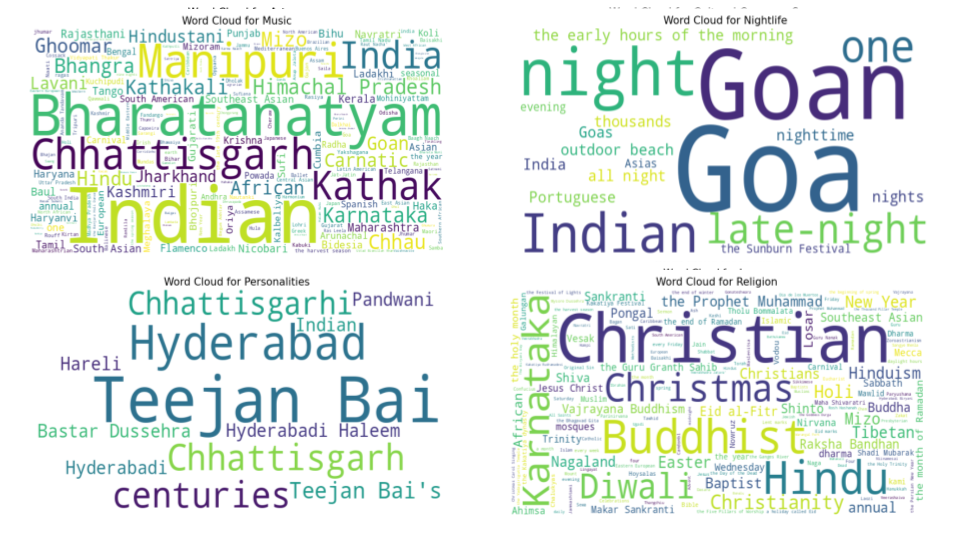}
    \caption{Word Cloud Representation based on attributes  Categories like transport, sports, tourism and ceremonies }
    \label{Word_Cloud_Representation_1}
\end{figure*}

\begin{figure*}
    \centering
    \includegraphics[width=0.95\linewidth]{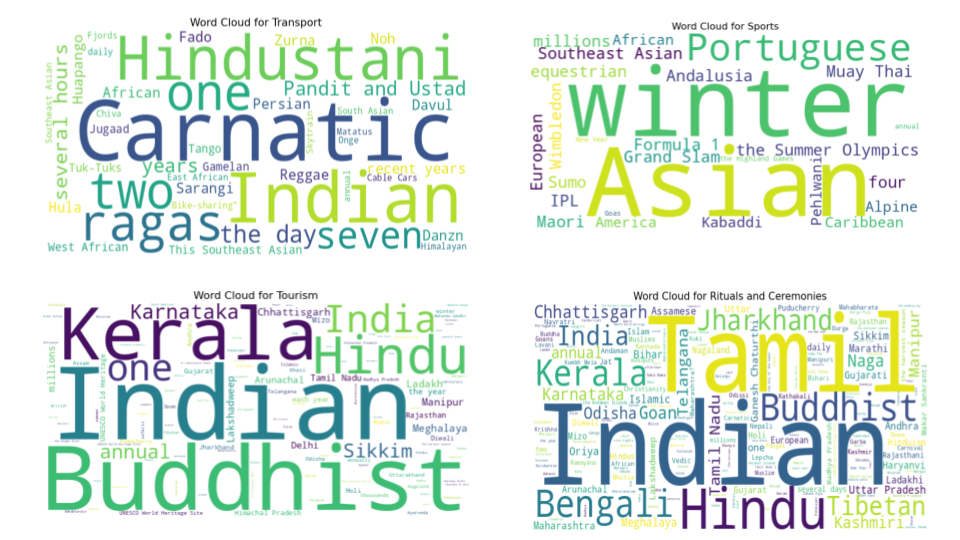}
    \caption{Word Cloud Representation based on attributes  Categories like history ,language ,medicine ,festivals.}
    \label{Word_Cloud_Representation_1}
\end{figure*}

\begin{figure*}[hbt!]
    \centering
    \begin{subfigure}[b]{0.48\textwidth}
        \centering
        \includegraphics[height=7cm, width=7cm]{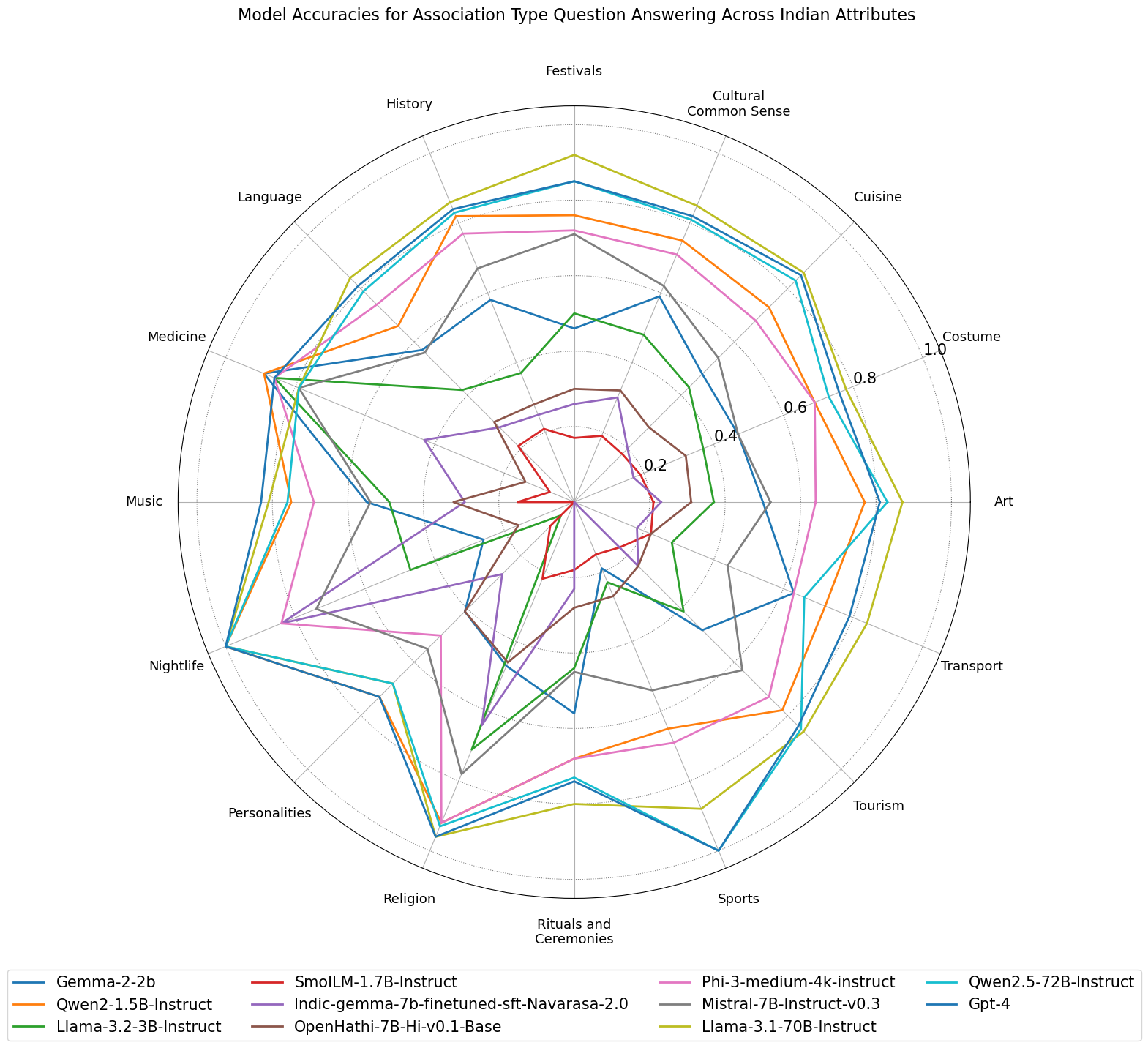}
        \caption{Performance of Language models on Association kind of Questions classified on the basis of attributes}
        \label{fig:association_attributes}
    \end{subfigure}
    \hfill
    \begin{subfigure}[b]{0.48\textwidth}
        \centering
        \includegraphics[height=7cm, width=7cm]{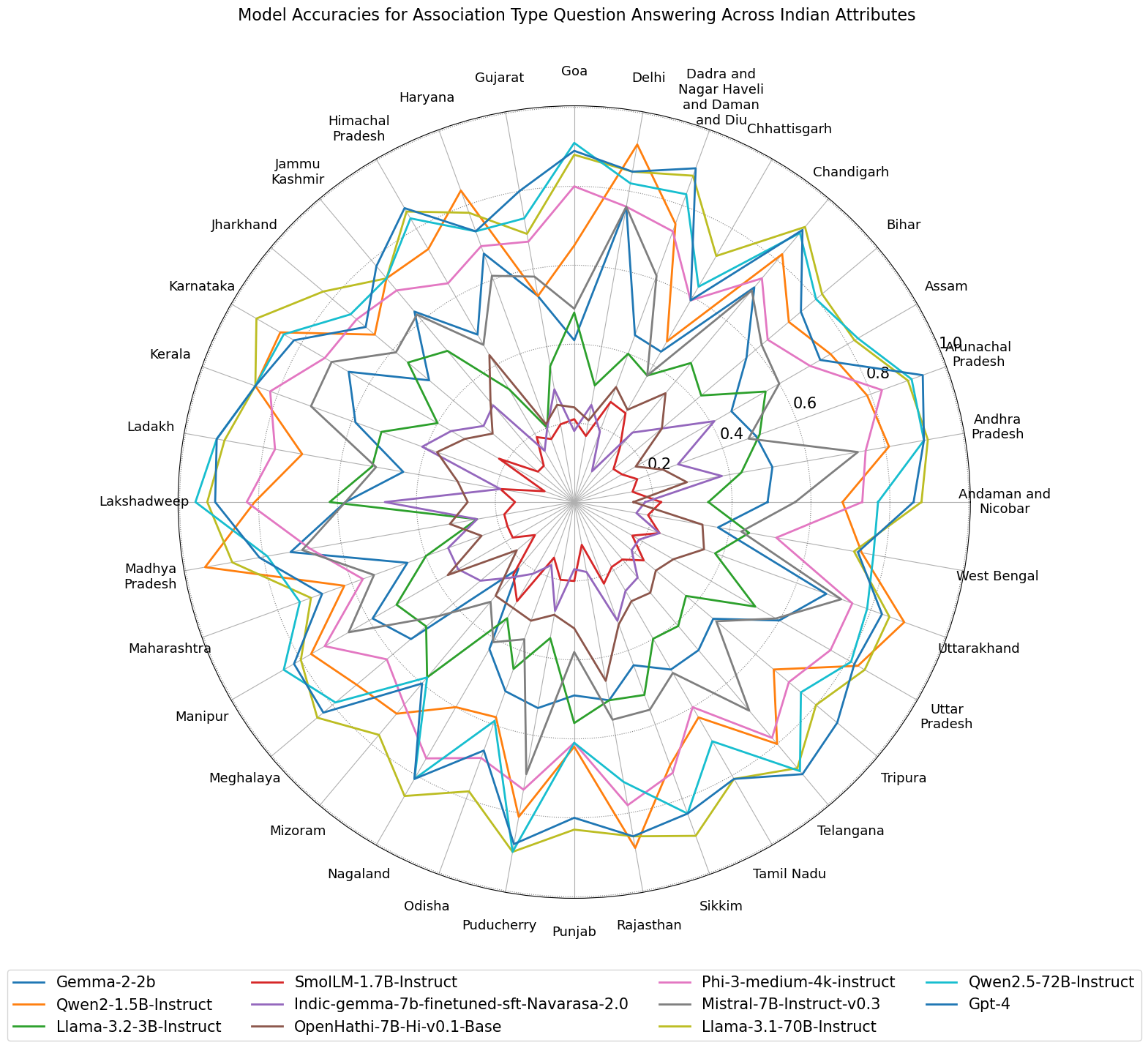}
        \caption{Performance of Language models on Association kind of Questions classified on the basis of states.}
        \label{fig:infographic_pipeline}
    \end{subfigure}
    \caption{Performance of language models on the  data points that falls under association kind of questions in \textit{\textbf{SANSKRITI}} that are further classified on the basis of attributes and states. } 
    \label{fig:association_states}
\end{figure*}

\begin{figure*}[hbt!]
    \centering
    \begin{subfigure}[b]{0.48\textwidth}
        \centering
        \includegraphics[height=7cm, width=7cm]{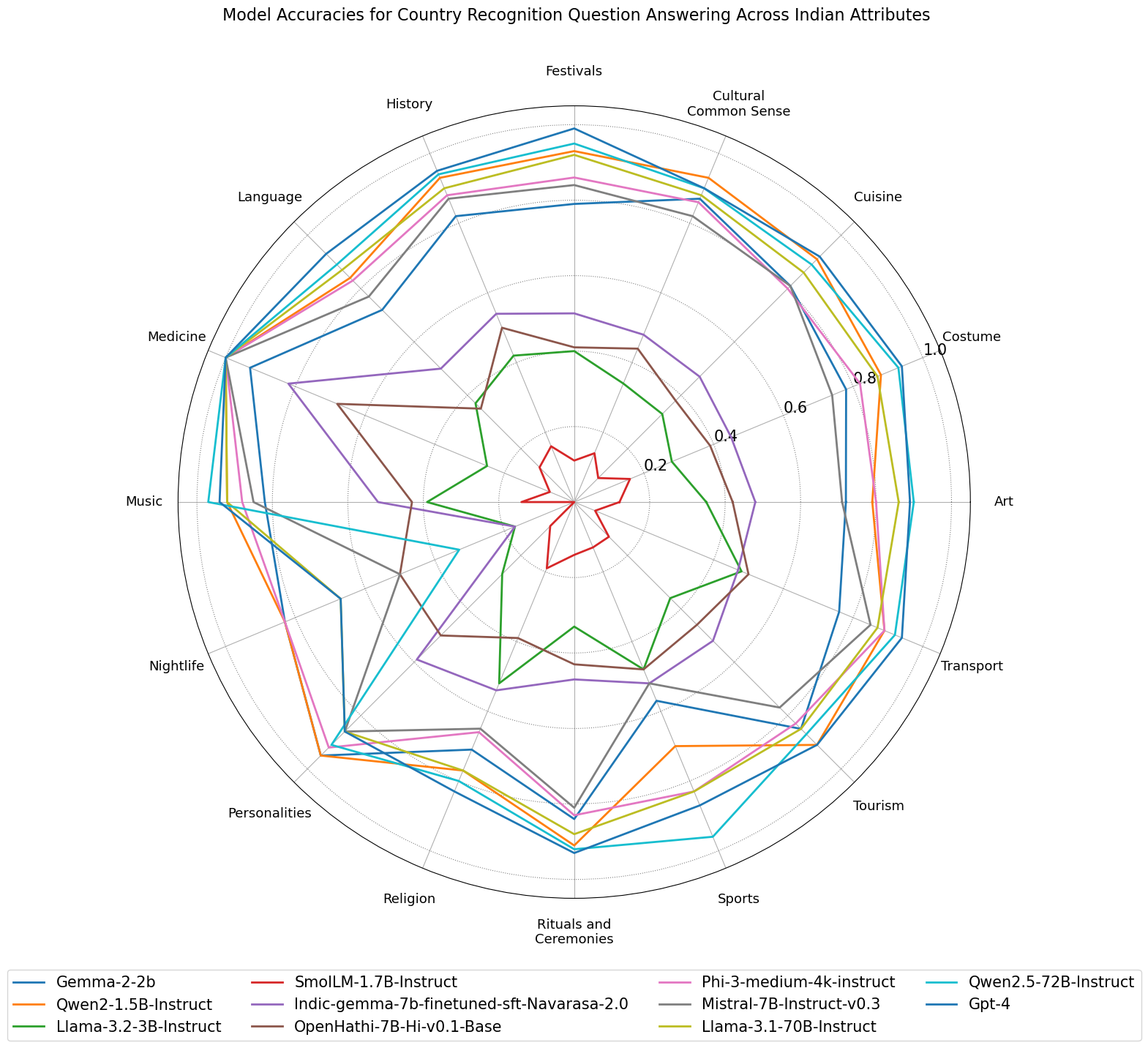}
        \caption{Performance of Language models on  Country\_Prediction kind of Questions classified on the basis of attributes}
        \label{fig:metadata_pipeline}
    \end{subfigure}
    \hfill
    \begin{subfigure}[b]{0.48\textwidth}
        \centering
        \includegraphics[height=7cm, width=7cm]{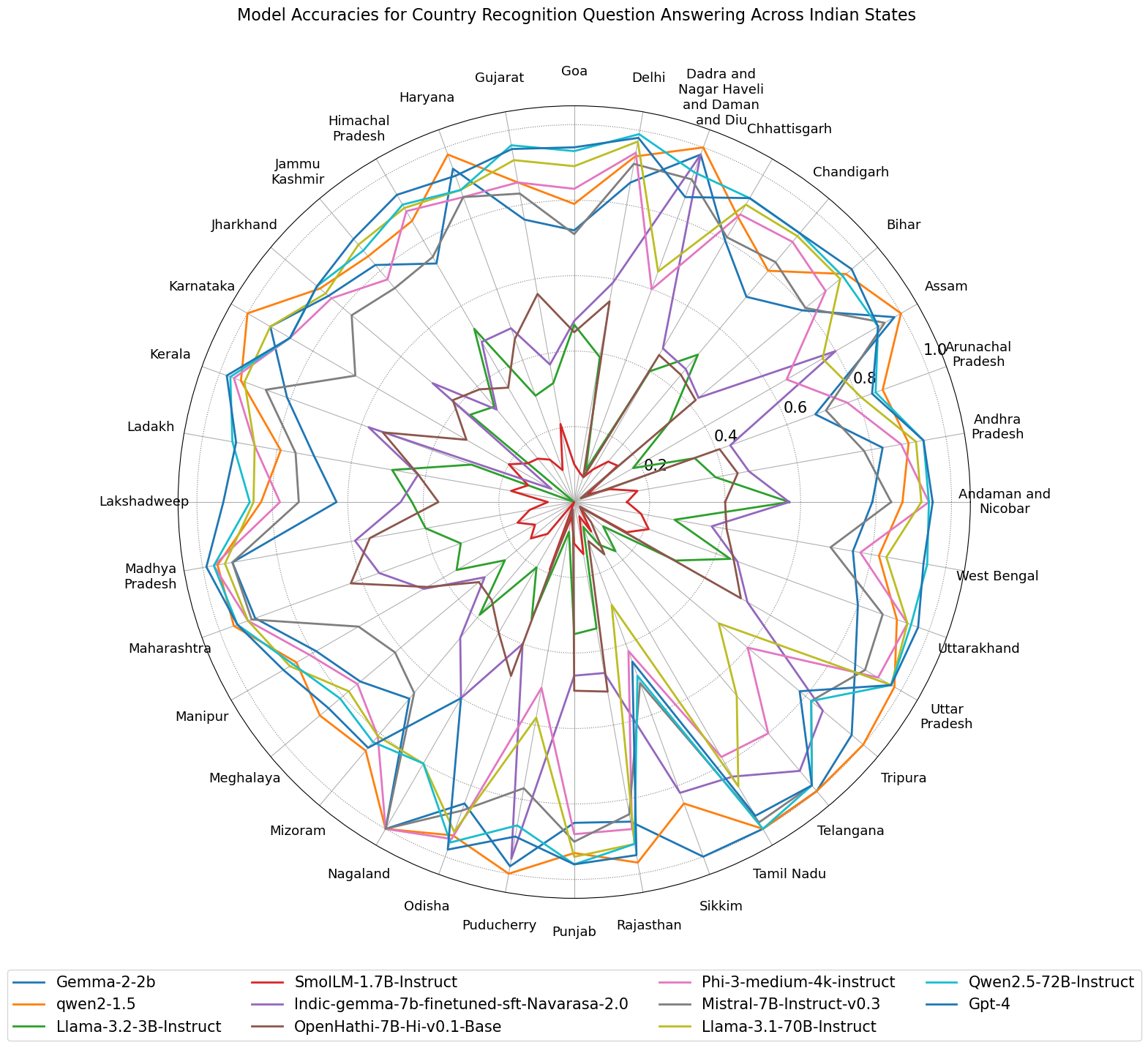}
        \caption{Performance of Language models on  Country\_Prediction kind of Questions classified on the basis of  states}
        \label{fig:infographic_pipeline}
    \end{subfigure}
    \caption{Performance of language models on the  data points that falls under country prediction kind of questions in \textit{\textbf{SANSKRITI}} that are further classified on the basis of attributes and states.} 
    \label{fig:data_gen_pipeline}
\end{figure*}
\begin{figure*}[hbt!]
    \centering
    \begin{subfigure}[b]{0.48\textwidth}
        \centering
        \includegraphics[height=7cm, width=7cm]{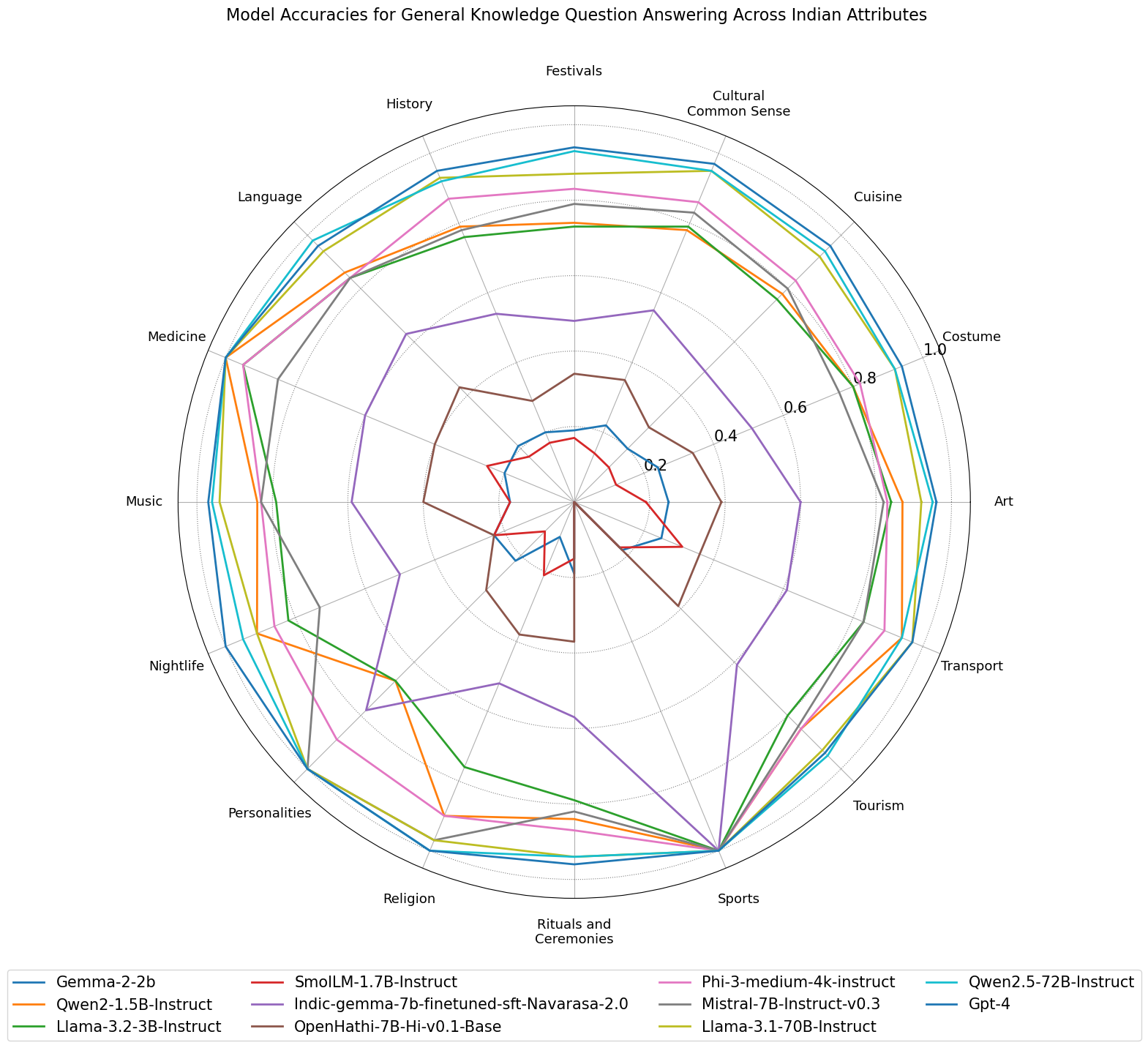}
        \caption{Performance of Language models on  General knowledge kind of Questions classified on the basis of  attributes.}
        \label{fig:metadata_pipeline}
    \end{subfigure}
    \hfill
    \begin{subfigure}[b]{0.48\textwidth}
        \centering
        \includegraphics[height=7cm, width=7cm]{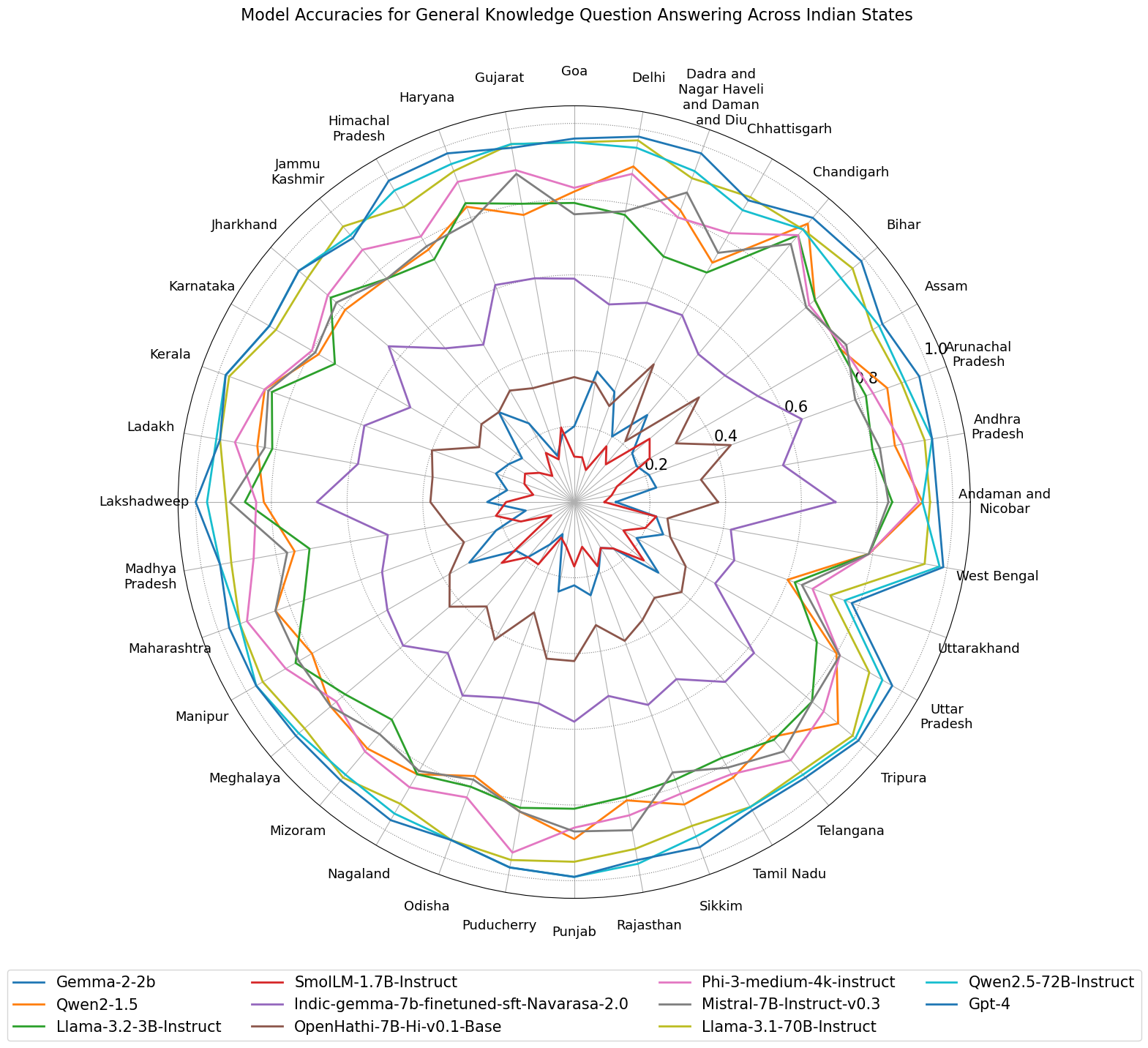}
        \caption{Performance of Language models on  General Knowledge  kind of Questions classified on the basis of  states.}
        \label{fig:infographic_pipeline}
    \end{subfigure}
    \caption{Performance of language models on the  data points that falls under General Knowledge  kind of questions in \textit{\textbf{SANSKRITI}} that are further classified on the basis of attributes and states.} 
    \label{fig:data_gen_pipeline}
\end{figure*}
\begin{figure*}[hbt!]
    \centering
    \begin{subfigure}[b]{0.48\textwidth}
        \centering
        \includegraphics[height=7cm, width=7cm]{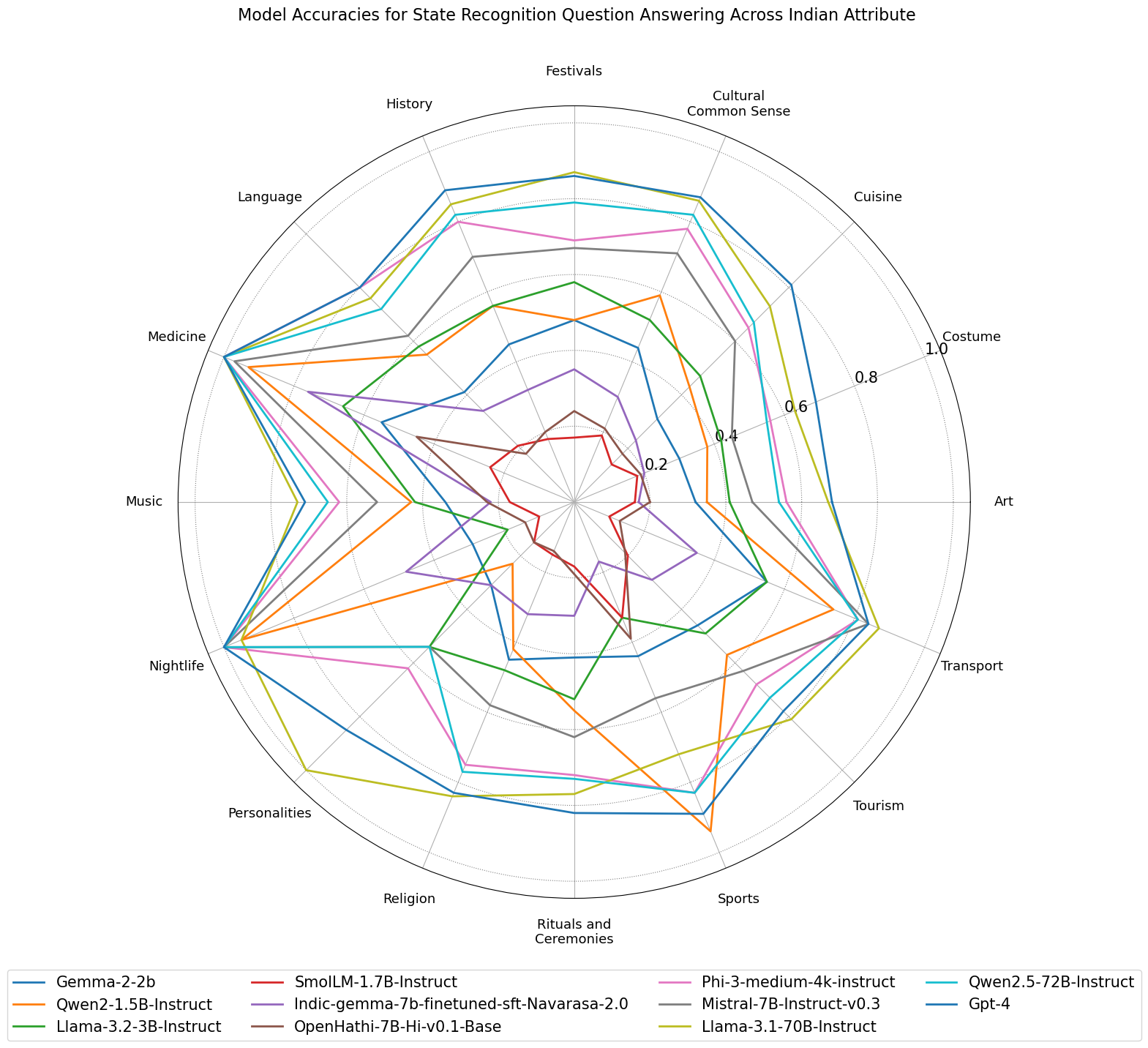}
        \caption{Performance of Language models on  State\_Prediction  kind of Questions classified on the basis of  attributes.}
        \label{fig:metadata_pipeline}
    \end{subfigure}
    \hfill
    \begin{subfigure}[b]{0.48\textwidth}
        \centering
        \includegraphics[height=7cm, width=7cm]{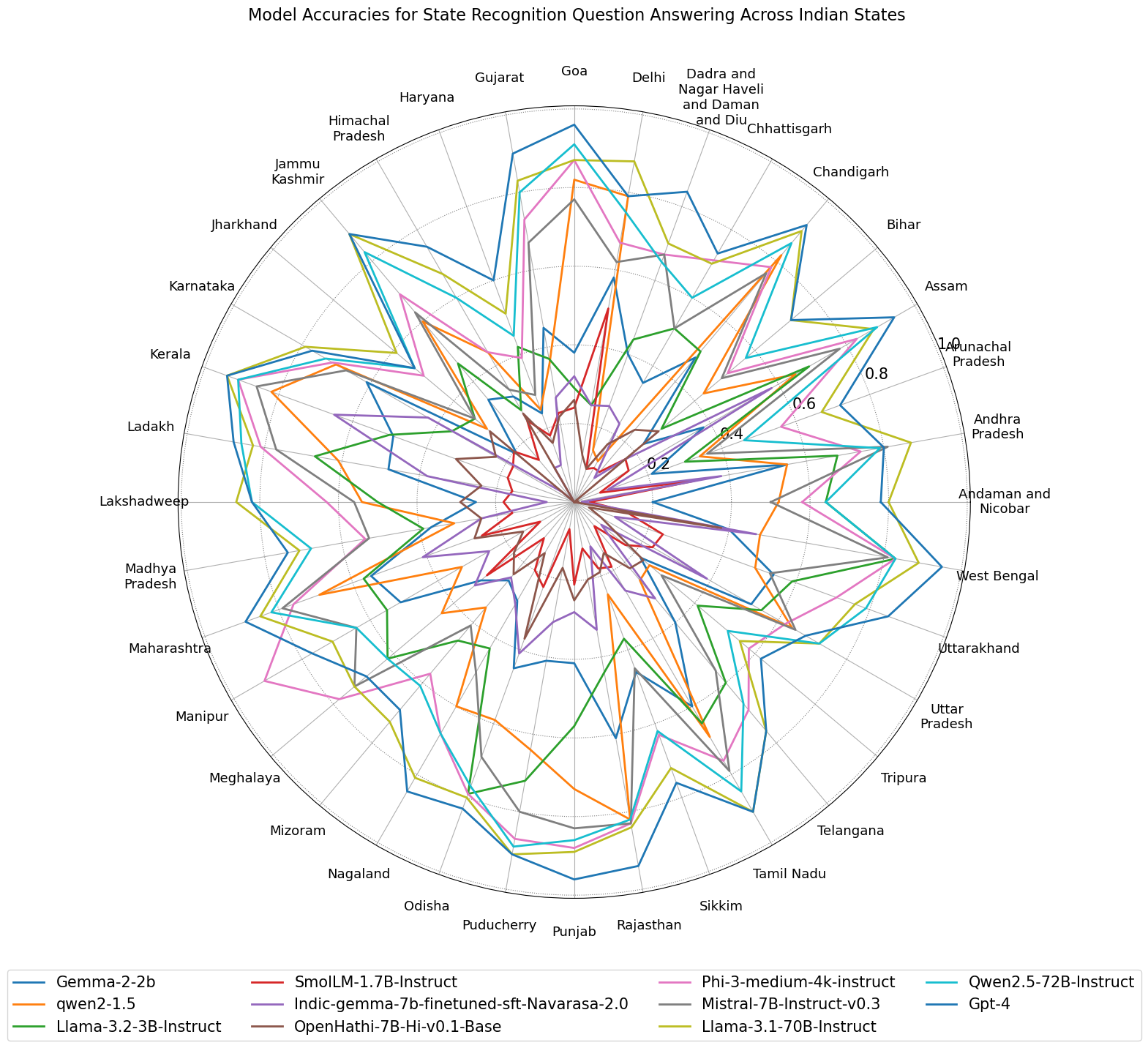}
        \caption{Performance of Language models on  State\_Prediction  kind of Questions classified on the basis of  states.}
        \label{fig:infographic_pipeline}
    \end{subfigure}
    \caption{Performance of language models on the  data points that falls under State\_Prediction  kind of questions in \textit{\textbf{SANSKRITI}} that are further classified on the basis of attributes and states.} 
    \label{fig:data_gen_pipeline}
\end{figure*}

\subsection{Question Formulation from the Source text and the Rationale behind it.}

\includepdf[pages=-,scale=0.85]{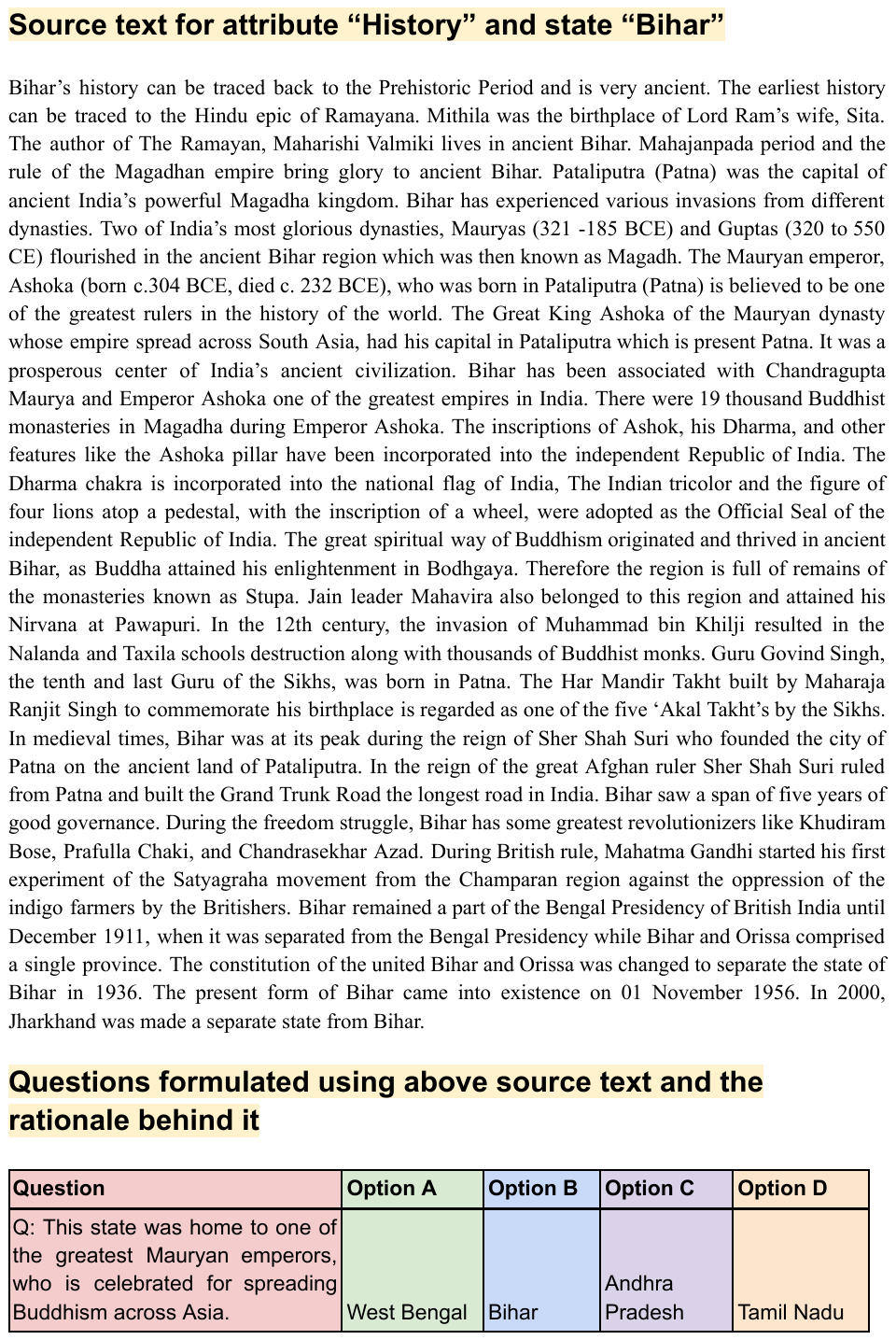}

\end{document}